\documentclass[twoside,11pt]{article}

\usepackage{jmlr2e}
\usepackage{lastpage}

\usepackage{microtype}

\usepackage[utf8]{inputenc}
\usepackage{amsmath,amsfonts}
\usepackage{amssymb}
\usepackage{graphicx}
\usepackage{tikz}
\usepackage{mathtools}
\usepackage{nccmath}
\usepackage{xspace}
\usepackage{xcolor}
\usepackage{booktabs}
\usepackage{array}
\usepackage{mathrsfs}
\definecolor{MyBlue}{rgb}{0.12, 0.25, 0.67}
\definecolor{MyPurple}{rgb}{0.8, 0.1, 0.8}
\definecolor{MyGreen}{rgb}{0.7, 0.1, 0.1}
\definecolor{MyRed}{rgb}{0.8, 0.1, 0.3}
\definecolor{MyYellow}{rgb}{0.7, 0.7, .1}
\definecolor{MyBlue2}{rgb}{0.1, 0.1, 0.6}
\definecolor{Color3}{HTML}{D62728}
\definecolor{Color1}{HTML}{0050C8}
\definecolor{Color2}{HTML}{4DAF4A}
\usepackage{pifont}

\usepackage[group-separator={,}]{siunitx}
\usepackage{url}
\usepackage{enumitem}
\usepackage{pdflscape}
\usepackage{verbatim}
\usepackage{arydshln}
\usepackage{multirow}
\usepackage{bigdelim}
\usepackage{stfloats}
\usepackage{pgfplots}
\usepackage{cleveref}
\newsavebox{\taxonomy}
\usepackage{makecell}

\usetikzlibrary{shadows,arrows,decorations,decorations.shapes,backgrounds,shapes,snakes,automata,fit,petri,shapes.multipart,calc,positioning,shapes.geometric,graphs,graphs.standard,plotmarks,math,arrows.meta}
\usepackage{tikz-qtree}

\newenvironment{restatedtheorem}[2]
  {\par\noindent\textbf{#1 #2.}\itshape}
  {\par}

\usepackage{algorithm}
\usepackage{algpseudocode}
\usepackage{algorithmicx}
\let\oldReturn\Return
\renewcommand{\Return}{\state\oldReturn}
\algtext*{EndWhile}
\algtext*{EndIf}
\algtext*{EndForAll}
\algtext*{EndFor}
\algtext*{EndFunction}

\usepackage{thmtools}
\usepackage{thm-restate}

\DeclareMathOperator*{\E}{\mathbb{E}}
\DeclareMathOperator*{\argmin}{arg\,min}
\DeclareMathOperator*{\len}{len}

\DeclareMathOperator*{\vol}{vol}
\DeclareMathOperator*{\diam}{diam}
\allowdisplaybreaks 

\newcommand\bbr{\mathbb{R}}
\newcommand\bbrpos{\mathbb{R}_{\ge 0}}
\newcommand\bbrspos{\mathbb{R}_{> 0}}
\newcommand\bbn{\mathbb{N}}
\newcommand\ep{\varepsilon}

\renewcommand{\d}[1]{\ensuremath{\operatorname{d}\!{#1}}}

\newcommand\hats{\hat{S}}
\newcommand\smola{\boldsymbol{a}}
\newcommand\smols{\boldsymbol{s}}
\newcommand\s{\mathcal{S}}
\newcommand\sm{\boldsymbol{s^m}}
\newcommand\A{\mathcal{A}}
\newcommand\bfone{\boldsymbol{1}}
\newcommand\norm[1]{||#1||}
\newcommand\tv[1]{||#1||_{TV}}
\newcommand\bfmu{\boldsymbol{\mu}}

\newcommand\M{\mathcal{M}}

\newcommand\tilpi{\tilde{\Pi}}
\newcommand\stilpi{\tilde{\pi}}
\newcommand\D{\mathcal{D}}

\newcommand\F{h} 
\newcommand\gk{\Gamma_k}
\newcommand\gff{\Gamma_{\textnormal{FF}}}
\newcommand\gac{\Gamma_{\textnormal{AC}}}

\newcommand\till{\tilde{\ell}}

\newcommand\scrf{\mathcal{H}} 
\newcommand\scrv[1]{\mathcal{V}(#1)} 
\newcommand\I{\mathcal{I}}
\newcommand\q{\boldsymbol{q}}
\newcommand\tilD{\tilde{\Delta}}
\newcommand\V{V} 
\newcommand\p{\mathcal{P}}
\newcommand\R{\mathcal{R}}
\newcommand\U{\boldsymbol{u}}
\newcommand\calU{\mathcal{U}}
\newcommand\rsa{\textnormal{Reg}_{\textnormal{SA}}}
\newcommand\rplus{\textnormal{Reg}_{\textnormal{AC}}^+}
\newcommand\rmul{\textnormal{Reg}_{\textnormal{AC}}^\times}
\newcommand\rmdp{\textnormal{Reg}_{\textnormal{MDP}}}

\newcommand\mum{\mu_0}

\jmlrheading{27}{2026}{1-\pageref{LastPage}}{9/25; Revised
4/26}{6/26}{25-2248}{Benjamin Plaut, Juan Li\'{e}vano-Karim, Hanlin Zhu, and Stuart Russell}

\ShortHeadings{Safe Learning Under Irreversible Dynamics via Asking for Help}{Plaut, Li\'{e}vano-Karim, Zhu, and Russell}
\firstpageno{1}

\begin{document}

\title{Safe Learning Under Irreversible Dynamics via Asking for Help\looseness=-1}

\author{\name Benjamin Plaut \email plaut@berkeley.edu 
\AND \name Juan Li\'{e}vano-Karim \email jp.lievano10@uniandes.edu.co
\AND \name Hanlin Zhu \email hanlinzhu@berkeley.edu
\AND \name Stuart Russell \email russell@berkeley.edu \vspace{.07 in}  \\ \addr Division of Computer Science\\ University of California\\ Berkeley, CA 94720-1776, USA
}

\editor{\textbf{Editor:} Laurent Orseau}

\maketitle

\begin{abstract}%
Most learning algorithms with formal regret guarantees essentially rely on trying all possible behaviors, which is problematic when some errors cannot be recovered from. Instead, we allow the learning agent to ask for help from a mentor and to transfer knowledge between similar states. We show that this combination enables the agent to learn both safely and effectively. Under standard online learning assumptions, we provide an algorithm whose regret and number of mentor queries are both sublinear in the time horizon for Markov decision processes with irreversible dynamics and infinite state spaces. Our proof involves a sequence of three reductions, making our result more general than a single algorithm. Conceptually, our result may be the first formal proof that it is possible for an agent to obtain high reward while becoming self-sufficient in an unknown, unbounded, and high-stakes environment without resets.\looseness=-1
\end{abstract}%


\section{Introduction}\label{sec:intro}

Concern has been mounting over a variety of risks from AI \citep{national_institute_of_standards_and_technology_us_artificial_2024,critch2023tasra,hendrycks2023overviewcatastrophicairisks,slattery2024ai}. Such risks include medical errors \citep{rajpurkar2022ai}, self-driving car crashes \citep{kohli2020enabling}, discriminatory sentencing \citep{villasenor2020artificial}, autonomous weapon accidents \citep{Abaimov2020}, bioterrorism \citep{mouton2024operational}, high-stakes cyberattacks \citep{guembe_emerging_2022}, and loss of control \citep{bengio2024managing}. These types of errors are particularly concerning because they are irreparable: a self-driving car cannot compensate for a crash by driving superbly later. We refer to irreparable errors as \emph{catastrophes}.\looseness=-1

%

Despite the gravity of these risks, there is little theoretical understanding of how to avoid them: nearly all of learning theory explicitly or implicitly assumes that any error can be recovered from. This assumption enables algorithms to cheerfully try all possible behaviors, since no matter how poorly the algorithm performs in the short term, it can always eventually make up for it. Although such approaches have produced positive results, we argue that they are fundamentally unsuitable for the high-stakes AI applications above: we do not want autonomous weapons or surgical robots to try all possible behaviors. One could train AI agents entirely in controlled lab settings where all errors \emph{are} recoverable, but we argue that sufficiently general agents will inevitably encounter novel scenarios when deployed in the real world. Models often behave unpredictably in unfamiliar situations (see, e.g., \citealp{quinonero2022dataset}), and we do not want AI biologists or self-driving cars to behave unpredictably.\looseness=-1

Instead, we suggest that agents should recognize when they are in unfamiliar situations and ask for help. When help is requested, a supervisor (here called a \emph{mentor} to reflect their collaborative role) guides the agent to avoid problematic actions without the agent needing to try those actions firsthand. This approach is particularly suitable for high-stakes applications where AI systems are already configured to work alongside humans. For example, imagine a human doctor who oversees AI surgeons and is prepared to take over in tricky situations, or a backup human driver in an autonomous vehicle. In these scenarios, occasional requests for human guidance are both practical and valuable for ensuring safety.

In order to avoid excessive requests for help, the agent must be able to accurately identify unfamiliar situations, i.e., situations that are not similar to any of the agent's prior experiences. We model ``similarity'' as a distance metric on the state space and assume that the agent can transfer knowledge between similar states, but this transfer becomes less reliable as the states become less similar. We call this assumption \emph{local generalization}.\footnote{See \Cref{sec:model-ac} for the formal definition.}


Guided by the principles above, we design an agent that (under standard online learning assumptions) performs nearly as well as the mentor while gradually becoming self-sufficient, even when some errors are irreparable.


\subsection{Our Model}\label{sec:model-intro}

We study online learning, where the agent must perform well while learning: finding a good policy is futile if irreparable damage is caused along the way. We allow the MDP to be \emph{non-communicating} (some states might not be reachable from other states) and do not allow the agent to be reset to the start state. This is the class of ``multichain MDPs'' \citep{puterman2014markov}.  Crucially, we allow an infinite state space: in the real world, an agent will likely never see the exact same state twice. Instead, the agent must use local generalization to judge how similar the present state is to previous states.



On each time step, the agent has the option to query a mentor. When queried, the mentor demonstrates the action they would take in the current state. We assume that the agent begins with no prior knowledge.\footnote{Future work could incorporate an initial offline data set. See \Cref{sec:conclusion} for further discussion.} We want to minimize \emph{regret}, defined as the difference between the expected sum of rewards obtained by the agent and the mentor when each is given the same starting state.  (Regret typically refers to the performance gap between the agent and an optimal policy, but we allow the mentor to be either optimal or suboptimal, so our definition is arguably more general.) An algorithm for this problem should satisfy two criteria:\looseness=-1
\begin{enumerate}[leftmargin=1.5em,
    topsep=0.5ex,
    partopsep=0pt,
    parsep=0pt,
    itemsep=0.5ex]
    \item The agent's performance should approach that of the mentor. Formally, the average regret per time step should go to 0 as the time horizon $T$ goes to infinity, or equivalently, the cumulative regret should be sublinear in $T$ (denoted $o(T)$). An algorithm that guarantees $o(T)$ regret is called a \emph{no-regret} algorithm.  
    \item The agent should become self-sufficient. Formally, the rate of querying the mentor should go to 0 as $T \to \infty$, or equivalently, the cumulative number of queries should be $o(T)$.
\end{enumerate}

\looseness=-1

\begin{figure}[h]
\footnotesize
\resizebox{\linewidth}{!}{%
\begin{tikzpicture}[
>={Stealth[scale=1.2]}, 
    semithick, 
    main/.style={rounded corners=3pt, align=center,inner xsep=4.5pt, inner ysep=7pt, path picture={\fill[left color=blue!8, right color=blue!2] (path picture bounding box.south west) rectangle (path picture bounding box.north east);}},
    edgenodelow/.style={align=center,midway,below=.04in},
    edgenodehi/.style={align=center,midway,above=.04in},
    arrow/.style={->}
]

\node[main] (mdp) at (0,0) {Online learning\\ in multichain MDPs\\with a mentor};

\node[main] (ac) at (4.8,0) {Avoiding\\ catastrophe in\\ online learning\\with a mentor};

\node[main] (active) at (9.2,0) {Standard active\\online learning};

\node[main] (ff-alg) at (13.5,0) {Standard\\full-feedback\\online learning};

\draw[arrow] (mdp) -- node[edgenodelow] {Sec. \ref{sec:mdp-to-ac}\\Thm. \ref{thm:mdp-to-ac}} node[edgenodehi] {Third\\reduction} (ac);

\draw[arrow] (ac) -- node[edgenodelow] {Sec. \ref{sec:ac-to-standard}\\Thm. \ref{thm:ac-to-standard}} node[edgenodehi] {Second\\reduction} (active);

\draw[arrow] (active) -- node[edgenodelow] {Sec. \ref{sec:active-to-full-feedback}\\Thm. \ref{thm:active-to-full}} node[edgenodehi] {First\\reduction} (ff-alg);

\end{tikzpicture}
}
\caption{An illustration of our three-step reduction.}
\label{fig:reduction}
\end{figure}

\subsection{Our Contribution}

We provide an algorithm which satisfies both criteria given local generalization and standard online learning assumptions. Our proof involves a three-step reduction across four models of online learning (\Cref{fig:reduction}).  Throughout the paper, we use ``catastrophe'' as a conceptual and informal term to refer to errors that immediately ruin the entire attempt; the precise reversibility assumptions (or lack thereof) are made clear when the models are formally defined later. Here we provide a brief conceptual overview of the models:\looseness=-1
\begin{enumerate}[leftmargin=1.5em,
    topsep=0.5ex,
    partopsep=0pt,
    parsep=0pt,
    itemsep=0.5ex]
    \item Standard full-feedback online learning, where catastrophe is impossible because the agent and the optimal policy are evaluated on the same states (and rewards are bounded).\looseness=-1
    \item Standard active online learning, which is the same as Model 1 but the agent only receives feedback when it queries.
    \item Avoiding catastrophe in online learning, a new model where catastrophe \emph{is} possible, but avoiding catastrophe is the agent's only goal (it does not also need to maximize reward).
    \item Online learning in multichain MDPs, where the agent must maximize reward (which requires avoiding catastrophe).
\end{enumerate}

 \Cref{tab:model-comparison} compares the models in more detail (note that Models 1 and 2 are combined under ``standard adversarial active online learning'').
 
\emph{First reduction.} We first show that any algorithm for Model 1 can be transformed into an algorithm for Model 2. This reduction is based on standard ideas from \citet{cesa2005minimizing}, who proved a similar result but only for a single algorithm. However, to our knowledge, our other two reductions are technically novel.

\emph{Second reduction.} The high-level idea is simple: mostly follow a ``default'' algorithm which works well in the absence of catastrophe, but ask for help in unfamiliar situations. However, bringing this to fruition requires carefully balancing the frequency of mentor queries, the frequency of errors, and the magnitude of errors.  Altogether, we show that if the ``default'' algorithm is effective for Model 2, then the overall algorithm is effective for Model 3. \looseness=-1


\emph{Third reduction.} Finally, we reduce Model 4 to Model 3. This reduction requires no modifications of the algorithm: we show that any algorithm which guarantees avoidance of catastrophe (in the sense of Model 3) also ensures high reward in multichain MDPs. This is \emph{not} saying that avoiding catastrophe in a particular MDP suffices for high reward in that MDP (which is generally false). This is saying that any algorithm guaranteed to avoid catastrophe in general must have certain properties which also guarantee high reward. The regret analysis involves a decomposition of regret into ``state-based'' and ``action-based'' regret which we believe is novel.\looseness=-1

Chaining together all three reductions produces (to our knowledge) the first no-regret guarantee for multichain MDPs with an infinite state space.

\begin{restatedtheorem}{Theorem}{\ref{thm:mdp-to-ac-final} \textnormal{(Informal)}}
Under standard online learning assumptions, there exists a no-regret algorithm for infinite-state multichain MDPs that makes $o(T)$ mentor queries.
\end{restatedtheorem}

\textbf{In other words, we prove that it is possible for an agent to obtain high reward while becoming self-sufficient in an unknown, unbounded, and high-stakes environment without resets.} We are not aware of any prior work that can be said to imply this conclusion.\looseness=-1

\section{Prior Work}\label{sec:related}

There is a vast body of work studying regret in MDPs, but all prior no-regret guarantees rely on restrictive assumptions. For context, the agent has no hope of success without further assumptions: any unexplored action could lead to paradise (so the agent cannot neglect it) but the action could also lead to disaster (so the agent cannot risk it). This intuition is formalized by the ``Heaven or Hell'' problem in \Cref{fig:heaven_hell_problem}. Our work is not immune to this problem: rather, we argue that access to a mentor is a natural and practical complement to typical assumptions like reversibility or periodic resets that may not always hold in high-stakes AI applications.\looseness=-1

\begin{figure}[b]
    \centering
    \footnotesize
    \begin{tikzpicture}[>={Stealth[scale=1]}, auto, node distance=3cm, semithick]
        \tikzstyle{state} = [ellipse, minimum width=1.6cm,minimum height=.6cm, inner sep=0,path picture={\fill[left color=blue!8, right color=blue!2] (path picture bounding box.south west) rectangle (path picture bounding box.north east);}]

        \node[state] (S1) {Start};
        \node[state, above right=0cm and 2cm of S1] (Heaven) {Heaven};
        \node[state, below right=0cm and 2cm of S1] (Hell) {Hell};

        \path[->] (S1) edge[bend left=12] node[midway, above=.1cm] {Action 1} (Heaven)
                       edge[bend right=12] node[midway, below=.1 cm] {Action 2} (Hell);
        \path[->] (Heaven) edge[loop right] node[right] {Reward = 1} (Heaven);
        \path[->] (Hell) edge[loop right] node[right] {Reward = 0} (Hell);

    \end{tikzpicture}
    \caption{The ``Heaven or Hell'' problem shows why online learning in MDPs is doomed without further assumptions. Action 1 leads to high reward forever and Action 2 leads to low reward forever. Without further assumptions, the agent has no way to tell which action leads to Heaven and which leads to Hell, so it is impossible to guarantee high reward.\looseness=-1}
    \label{fig:heaven_hell_problem}
\end{figure}

\Cref{fig:prior_work} provides a taxonomy of assumptions used to handle the Heaven or Hell problem, which will guide our discussion of related work. Unless otherwise specified, citations are representative, not exhaustive.

\begin{lrbox}{\taxonomy}
\begin{tikzpicture}[>={Stealth[scale=1]},
  level 1/.style={sibling distance=48mm},
  level 2/.style={sibling distance=23mm},
  level 3/.style={sibling distance=28mm},
  every node/.style={align=center, rectangle, rounded corners=5pt, font=\footnotesize, align=center, text centered, inner xsep=4pt, inner ysep=6pt},
  level distance=17mm,
    edge from parent/.style={draw,->,black!80},
    edge from parent path={
            (\tikzparentnode\tikzparentanchor) edge  (\tikzchildnode\tikzchildanchor)
    },
    approach/.style={
        path picture={\fill[left color=blue!8, right color=blue!2] (path picture bounding box.south west) rectangle (path picture bounding box.north east);}
    },
    subnote/.style={
path picture={\fill[left color=blue!8, right color=blue!2] (path picture bounding box.south west) rectangle (path picture bounding box.north east);}
    }
]

\newcommand\citeplaceholder[1]{}


\node[approach] {Assumptions powering\\ online learning in MDPs}
  child {node[approach] {All errors\\are recoverable}
    child {node[approach] {All actions\\are reversible }
    	child {node[subnote] {Reset after\\each episode \citeplaceholder{azar_minimax_2017}}}
	child {node[subnote] {Communicating\\MDP      \citeplaceholder{jaksch_near-optimal_2010}}}
    }
    child {node[subnote] {Regret is\\relative to\\prior errors\citeplaceholder{he2021nearly, lattimore_asymptotically_2011, liu_regret_2021}}}
  }
  child {node[approach] {Prior\\knowledge}
    child {node[subnote] {Safety\\constraint\\is known \citeplaceholder{zhao_state-wise_2023, model_zhao_22a}}}
    child {node[subnote] {Strictly safe\\ policy\\is known \citeplaceholder{liu2021learning,stradi2024learning}}}
  }
  child {node[approach] {External\\help}
    child {node[approach] {Focus is\\safety and\\reward}
        child {node[subnote,        path picture={\fill[left color=green!9, right color=green!3] (path picture bounding box.south west) rectangle (path picture bounding box.north east);}] {\bf Our work:\\\bf no-regret\\\bf guarantee}}
        child {node[subnote] {Other measures\\of performance\citeplaceholder{kosoy_delegative_2019, maillard_active_2019}}}
    }
    child {node[subnote] {Focus is\\safety only  \citeplaceholder{cohen_pessimism_2020,plaut_avoiding_2024}}}
  };
\end{tikzpicture}
\end{lrbox}

\begin{figure}[h!]
\usebox{\taxonomy}
\caption{A taxonomy of assumptions which enable meaningful theoretical results for online learning in MDPs. \Cref{fig:heaven_hell_problem} shows why meaningful guarantees are impossible without further assumptions.}
\label{fig:prior_work}
\end{figure}

\subsection{Assuming That All Errors Are Recoverable} 

The most common approach is to assume that any error can be offset by performing well enough in the future. This assumption can take various forms, including periodically resetting the agent \citep{azar_minimax_2017} or assuming a communicating MDP \citep{jaksch_near-optimal_2010}. A less obvious form is defining regret relative to prior errors, i.e., allowing catastrophic errors as long as the agent does as well as possible going forward. This approach compares the performance of the agent and the reference policy \emph{on the agent's state sequence} and does not consider whether the reference policy would have led to better states. For example, an agent which enters Hell in \Cref{fig:heaven_hell_problem} is considered successful, since all actions are equivalent thereafter and thus it obtains ``optimal'' reward on all future time steps. This comparison can be formulated as a type of regret \citep{he2021nearly, liu_regret_2021} or via the related notion of ``asymptotic optimality'' due to \citet{lattimore_asymptotically_2011}.\looseness=-1

Regardless of the specific form of the assumption, this approach faces difficulty in high-stakes domains where some errors cannot be recovered from. For example, a robotic vehicle cannot make up for a crash by driving superbly afterward. Thus we argue that this approach is ill-suited for the types of AI risks we are concerned with.

One notable paper in this category is \citet{ortner2012online}, which studies infinite-state MDPs with the recoverability assumption and a Lipschitz assumption similar to local generalization. Their regret bound, like ours, has exponential dependence on the dimension of the metric space (in fact, with the exact same exponent of $\frac{2n+1}{2n+2}$).

\subsection{Prior Knowledge} 

A smaller body of work allows for the possibility of catastrophe but provides the agent sufficient prior knowledge to avoid it, primarily within the formalism of constrained MDPs (CMDPs) \citep{altman1999constrained}. Some work assumes that the exact safety constraint is known, i.e., the agent knows upfront exactly which actions cause catastrophe \citep{model_zhao_22a, zhao_state-wise_2023}. Other work assumes the agent knows a fully safe policy and the agent is periodically reset \citep{liu2021learning,stradi2024learning}. In contrast, we assume no prior knowledge of the environment.\looseness=-1

\subsection{External Help} An even less common approach is to allow the agent to ask for help from a mentor. Crucially, this approach had not previously produced a no-regret guarantee for large or infinite multichain MDPs. Most prior work in this category focuses on safety alone, and if reward guarantees are provided, they fall into the ``defining regret relative to prior errors'' category \citep{cohen_pessimism_2020,cohen_curiosity_2021}. The paper most relevant to our work is \citet{plaut2025avoiding}, who proposed the ``avoiding catastrophe in online learning with a mentor'' model referred to in \Cref{fig:reduction}. \Cref{thm:ac-to-standard-final} recovers the primary result of that paper but with a more general proof that applies to any algorithm which meets the reduction criteria. In contrast, that paper's result only applies to a specific algorithm. More importantly, that paper only considers avoiding catastrophe and does not consider any effectiveness-related reward. In contrast, \Cref{thm:ac-to-standard-final} is only a stepping stone to our main result: a guarantee of high reward in multichain MDPs (which requires both effectiveness and avoiding catastrophe).


We are aware of just two papers that use external help to obtain regret guarantees on reward in multichain MDPs: \citet{maillard_active_2019} and \citet{kosoy_delegative_2019}. First and foremost, both papers assume a finite state space, which means that with a large enough time horizon, the agent can eventually query every state. In contrast, in the real world, the exact same state never appears twice. Furthermore, both papers' regret bounds scale linearly with the number of states. Additional differences include:

\begin{enumerate}[leftmargin=1.5em,
    topsep=0.5ex,
    partopsep=0pt,
    parsep=0pt,
    itemsep=0.5ex]
    \item The regret bound of \citet{maillard_active_2019} includes an instance-dependent $\gamma^*$ parameter, whereas our bound is worst-case across all multichain MDPs that satisfy our assumptions.
    \item The method of \citet{kosoy_delegative_2019} requires computing a Bayesian posterior, which is intractable without strong assumptions. In their case, they assume a finite number of possible hypotheses.
    \item Like us, \citet{kosoy_delegative_2019} allows a suboptimal mentor. Unlike us, they show that the agent can still obtain asymptotically optimal reward as long as the mentor only takes safe actions and has a nonzero probability of taking the optimal action.
    \item \citet{kosoy_delegative_2019} studies the infinite-horizon setting where ``no-regret'' means that the average regret goes to 0 as the discount factor goes to 1.
    \item By assuming a finite state space, both \citet{maillard_active_2019} and \citet{kosoy_delegative_2019} avoid the need for a similarity metric between states. As a result, they avoid the exponential dimensionality dependence in the regret bound that is typical for online learning in metric spaces (e.g., our work and \citealp{ortner2012online}).
\end{enumerate}

\subsection{Other Related Work} There is a wide range of other work on designing agents to act cautiously when uncertain. Representative theoretical papers include \citet{cortes2018online, hadfield-menell_inverse_2017, li2008knows}. Our model is also reminiscent of active learning \citep{hanneke2014active} due to the agent proactively asking for help and of imitation learning \citep{osa_algorithmic_2018} due to the agent attempting to follow the mentor policy. However, we are not aware of any direct technical implications. Finally, see \citet{garcia_comprehensive_2015,gu_review_2024, krasowski23} for surveys of the safe reinforcement learning literature and \citet{cesa2006prediction} for a survey of online learning in general.

\section{Preliminaries}\label{sec:model}

Here we describe the general online learning setup which is shared by all models we study. Each specific model is defined in its relevant section (Sections \ref{sec:model-standard}, \ref{sec:model-ac}, and \ref{sec:model-mdp}). As such, the setup here is underspecified and also more general than each specific model. In particular, the adversary framework here generalizes MDPs, which are not defined until \Cref{sec:model-mdp}. \looseness=-1



\subsection{States, Actions, Queries, and Histories} Let $\Delta(X)$ denote the set of probability distributions on a set $X$, let $\bbn$ denote the strictly positive integers, and let $[k] = \{1,\dots,k\}$ for $k \in \bbn$. Let $\s$ be a state space, $\A$ be a finite action space, and $T \in \bbn$ be a time horizon. We assume that $\s$ is countable for mathematical convenience, but there are no fundamental obstacles for continuous spaces. On each time step $t \in [T]$, the agent observes a state $s_t \in \s$ and selects an action $a_t \in \A$. Let $\smols = (s_1,\dots,s_T)$ and $\smola = (a_1,\dots,a_T)$.\looseness=-1

Each time step $t$ is also associated with a ``correct'' action $a_t^m$ which can be interpreted as the mentor's action in state $s_t$ (although we keep $a_t^m$ more general for now). In addition to choosing an action $a_t$, the agent also makes a query decision $q_t \in \{0,1\}$. The agent observes $a_t^m$ if and only if $q_t = 1$. With slight abuse of notation, we can express this concisely by assuming that the agent observes $a_t^m q_t$: if the algorithm queries, then $q_t = 1$ so $a_t^m = a_t^m q_t$ is observed. If the algorithm does not query, then $0  = a_t^m q_t$ is observed. (We assume 0 is a  ``dummy'' action not in $\A$.) In general we assume that the agent observes $a_t^m q_t$ before selecting $a_t$, but we also consider ``query-agnostic'' algorithms (see below).\looseness=-1

For $t \in [T+1]$, let $\F_t = (s_i, a_i, a_i^m q_i)_{i \in [t-1]}$ be the history at the end of time step $t-1$ (or equivalently, the start of time step $t$).\footnote{Since there is technically no time step 0 or $T+1$, $\F_1$ and $\F_{T+1}$ must be interpreted as the history at the start of time step 1 and the end of time step $T$, respectively.} Let $\scrf = (\s \times \A \times(\A\cup\{0\}))^*$ be the set of possible histories. Note that $a_i^m q_i$ also tells us the value of $q_i$. We use $\smallfrown$ to denote vector concatenation: for example, $\F_{t+1} = \F_t \smallfrown (s_t, a_t, a_t^m q_t)$.

\subsection{Adversaries and Algorithms} On each time step $t$, the state $s_t$ and correct action $a_t^m$ are determined by the \emph{adversary}. Formally, an adversary $\V$ is a function $\V: \scrf \to \Delta(\s \times \A)$ which takes as input $\F_t$ and returns the distribution of $s_t$ and $a_t^m$. In general, $\V$ can be any function, although for MDPs the adversary must use a Markov transition function.\looseness=-1

The action $a_t$ and query decision $q_t$ are determined by the algorithm. Formally, an \emph{active learning algorithm} (or just ``algorithm'') $\Gamma$ consists of a query function $\Gamma^Q: \scrf \times \s \to \Delta(\{0,1\})$ and an action function $\Gamma^\A: \scrf \times \s \times (\A\cup\{0\}) \to \Delta(\A)$. The query function takes as input $\F_t$ and $s_t$ and returns the distribution of $q_t$. The action function takes as input $\F_t,s_t,$ and the query result $a_t^mq_t$ and returns the distribution of $a_t$. Together, $\Gamma$ and $\V$ determine the distributions of $\smols,\smola,$ and $\q$. Unless otherwise stated, all expectations are over the randomness of $\smols,\smola,$ and $\q$. We assume $T$ is known, so $\V,\Gamma^Q,$ and $\Gamma^{\A}$ can all depend on it.\looseness=-1

An algorithm $\Gamma$ is \emph{full-feedback} if it always queries, i.e., $\Gamma^Q(\F, s) = 1$ for all $T\in \bbn,\F \in \scrf,s \in \s$.\footnote{In typical online learning, observing $a_t^m$ suffices to compute the loss of every action. See \Cref{sec:model-standard}.} An algorithm $\Gamma$ is \emph{query-agnostic} if its action does not depend on the query result from that round, i.e., $\Gamma^\A(\F,s,0) = \Gamma^\A(\F,s,a)$ for any $\F \in \scrf, s \in \s, a \in \A$. Equivalently, query-agnostic algorithms only observe the query result $a_t^m q_t$ after choosing $a_t$. For query-agnostic algorithms, we write $\Gamma^{\A}(\F_t,s_t)$ for simplicity. Full-feedback query-agnostic algorithms are the default in the online prediction literature and are typically defined without involving queries at all, but we treat this as a special case of our broader active learning model.\looseness=-1

\subsection{Overall Learning Protocol} To summarize, each time step $t \in [T]$ proceeds as follows:
\begin{enumerate}[leftmargin=1.5em,
    topsep=0.5ex,
    partopsep=0pt,
    parsep=0pt,
    itemsep=0.5ex]
    \item The adversary selects a state and correct action $(s_t,a_t^m) \sim \V(\F_t)$.
    \item The algorithm observes $s_t$ and selects a query decision $q_t \sim \Gamma^Q(\F_t,s_t)$.
    \item The algorithm observes $a_t^m q_t$ and selects an action $a_t \sim \Gamma^\A(\F_t, s_t, a_t^m q_t)$.

\end{enumerate}

\subsection{Objectives} The agent has two objectives: the number of queries and the regret. These are separate objectives and the agent must perform well according to both.\footnote{One could combine these into a single objective, but it is unclear how to choose the tradeoff parameter.\looseness=-1} Let $Q(T,\Gamma,\V) = \E[\sum_{t=1}^T q_t]$ be the expected number of queries. We want $Q(T,\Gamma,\V)$ to be sublinear in $T$. The regret objective varies between the models, but all versions involve optimizing over a known \emph{policy} class $\Pi$. Each (deterministic) policy $\pi \in \Pi$ is a function $\pi: \s \to \A$. \looseness=-1 

\begin{table}
\small
\centering
\resizebox{\linewidth}{!}{%
\hspace{-.1 in}
\begin{tabular}{l l l l}
& \textbf{Standard adversarial} & \textbf{Avoiding catastrophe} & \textbf{Multichain MDPs} \\
\midrule
Conceptual goal & \makecell[l]{High reward (assume\\no catastrophe)} & Avoid catastrophe only & \makecell[l]{High reward (requires\\avoiding catastrophe)} \\
\midrule
Desired regret & Sublinear in $T$ & Subconstant in $T$ & Sublinear in $T$\\
\midrule
Regret definition & \makecell[l]{Single state\\ sequence $\smols$} & \makecell[l]{Single state\\ sequence $\smols$} & \makecell[l]{Separate $\smols,\sm$ for\\ agent and mentor}\\
\midrule
Source of each state & Adversary & Adversary & Transition function \\
\midrule
Local generalization & No & Yes & Yes\\
\midrule
\makecell[l]{When are query\\ results observed?} & After choosing action & Before choosing action & Before choosing action\\
\bottomrule
\end{tabular}
}
\caption{The differences between the online learning models we consider. There is no column for full-feedback online learning, since it is a special case of (standard adversarial) active online learning.}
\label{tab:model-comparison}
\end{table}

\subsection{Further Technical Definitions} These definitions are not necessary to grasp the key ideas of the paper but are necessary to understand the precise technical results. For random variables $X$ and $Y$, let $\D(X)$ and $\D(X\mid Y)$ respectively be the distribution of $X$ and the conditional distribution of $X$ with respect to $Y$. For a distribution $p$, let $p(S)$ be the probability that $p$ assigns to a set $S$. For example, for a random variable $X$, $\D(X)(S) = \Pr[X \in S]$. When $S$ contains a single element $s$, we write $\D(X)(\{s\}) = \D(X)(s)$ for simplicity. For $c \in [0,1]$, let $\text{Bernoulli}(c)$ be the random variable which is 1 with probability $c$ and 0 otherwise.\looseness=-1
 
We consider both full adversaries and ``smooth'' adversaries which are required to choose a not-too-concentrated distribution over the next state. This allows a smooth interpolation between the ``easy'' case of i.i.d. states and the ``hard'' case of fully adversarial states. Given two distributions $p,\beta$ on $\s$, we say that $p$ is \emph{$\sigma$-smooth} (with respect to $\beta$) if $p(X) \le \frac{1}{\sigma}\beta(X)$ for every $X\subseteq\s$. The uniform distribution is a common choice for the ``baseline'' distribution $\beta$, but our results hold for any fixed $\beta$. Throughout the paper, we assume an arbitrary but fixed $\beta$. For convenience, we say that every distribution on $\s$ is $0$-smooth; this corresponds to the fully adversarial case. For $\sigma \in [0,1]$, an adversary $\V$  is $\sigma$-smooth if for any $\F \in \scrf$, the marginal distribution of $\V(\F)$ on $\s$ is $\sigma$-smooth. Equivalently, $\Pr_{(s,a)\sim \V(\F)} [s \in X] \le \frac{1}{\sigma}\beta(X)$ for any $X\subseteq\s$. Let $\scrv{\sigma}$ denote the set of $\sigma$-smooth adversaries. Since we will often handle $\sigma = 0$ separately, define the (scalar) Moore-Penrose pseudoinverse $\sigma^+$ by $\sigma^+ = \sigma^{-1}$ for $\sigma > 0$ and $0^+ = 0$. For convenience, we treat $\min_{x \in \emptyset} f(x)$ as $\infty$ for any function $f$.\looseness=-1

\section{First Reduction: Standard Adversarial Active Learning $\to$ Full-Feedback\looseness=-1}\label{sec:active-to-full-feedback}

We now present our first reduction (recall the roadmap in \Cref{fig:reduction}). We show that in the standard adversarial online learning model, any full-feedback algorithm can be converted into an active learning algorithm, where the regret of the latter is bounded by the regret of the former and the expected number of queries.

\subsection{The Standard Adversarial Online Learning Model}\label{sec:model-standard}

The agent's goal in this model is to predict the correct action (this is sometimes called the ``online prediction'' model). Formally, the objective in this model is to minimize a known loss function $\ell: \A \times \A \to [0,1]$, where $\ell(a,a^m)$ is the loss from predicting $a$ when the correct action is $a^m$. (In this model, all dependency on the states is captured implicitly through the actions.) One particular $\ell$ of interest is the binary loss $\ell_{\bfone}(a,a^m) = \bfone(a\ne a^m)$. Since $\ell$ is known, a full-feedback algorithm will observe $a_t^m$ on every time step and thus can compute the loss of every action in $\A$. The standard adversarial regret of an algorithm $\Gamma$ given an adversary $\V$  is\looseness=-1
\[
\rsa(T, \Gamma, \V,\ell) = \sup_{\pi \in \Pi}\, \E\left[ \sum_{t=1}^T \ell(a_t, a_t^m) - \sum_{t=1}^T \ell(\pi(s_t), a_t^m)\right].
\]
The typical goal is for $\rsa$ to be sublinear in $T$, or equivalently, for the average regret per time step to go to 0. We assume throughout this section that all algorithms are query-agnostic, i.e., $a_t^m q_t$ cannot be used to choose $a_t$ (equivalently, $a_t^mq_t$ is observed after $a_t$ is chosen). This aligns our model with the standard adversarial online learning model in the literature.

\subsection{The Reduction}

To define our algorithm, some additional notation is needed. For a history $\F$, define the length of $\F$, denoted $\len(\F)$, to be the total number of time steps covered by $\F$. Note that $\len(\F_t) = t-1$. Given a history $\F = (s_i', a_i', a_i^{m\prime} q_i')_{i \le \len(\F)}$ and vector $\U \in \{0,1\}^T$, define the query-restricted history $\F \cap \U = (s_i', a_i', a_i^{m\prime} q_i')_{i \le \len(\F): u_i = 1}$. We will sometimes apply these definitions with $\U \ne \q$ and $(s_i',a_i',a_i^{m'},q_i') \ne (s_i,a_i,a_i^{m},q_i)$.

Given a full-feedback algorithm $\gff$ and expected query budget $k$ (which need not be an integer), we define a new algorithm $\gk$ by running $\gff$ on the query-restricted history and querying independently with probability $k/T$ on each time step. Formally,
\begin{align*}
\gk^Q(\F,s) =&\ \text{Bernoulli}(k/T) \text{ and}\\
\gk^{\A}(\F,s) =&\ \gff^{\A}(\F\cap (q_1, \dots,q_{\len(\F)}),s).
\end{align*}
We call attention to the following aspects of this definition:
\begin{enumerate}[leftmargin=1.5em,
    topsep=0.5ex,
    partopsep=0pt,
    parsep=0pt,
    itemsep=0.5ex]
    \item If the algorithm is given a history $\F$, then $\len(\F)$ time steps have already occurred and so the first $\len(\F)$ query decisions must be known to the algorithm. This enables the algorithm to compute $\F\cap (q_1, \dots,q_{\len(\F)})$. Furthermore, we have $\F\cap (q_1, \dots,q_{\len(\F)}) = \F\cap \q$: by the definition of a query-restricted history, only time steps $t \le \len(\F)$ can appear in $\F\cap\q$, so the variables $(q_{\len(\F)+1},\dots,q_T)$ are ignored. 
    \item Since $\gk^{\A}$ does not depend on the query from the current time step, it is query-agnostic.
    \item The algorithm as written needs to know $T$, but this requirement can be removed by standard doubling-trick arguments (\citealp{besson2018doubling}; Chapter 2.3 in \citealp{cesa2006prediction}).
\end{enumerate}
The algorithm $\gk$ can equivalently be defined by the pseudocode in Algorithm~\ref{alg:active-to-full}.


\begin{algorithm}
\begin{algorithmic}[1]
\State Inputs: {full-feedback algorithm $\gff,\, T \in \bbn,\, k \in (0,T]$}
\For{$t$ \textbf{from} $1$ \textbf{to} $T$}
\State $\tilde{\F}_t \gets (s_i,a_i,a_i^m)_{i \in [t-1]:q_i = 1}$ \Comment{Equal to $\F_t \cap \q$}
    \State Observe $s_t$ from adversary
    \State Sample $a_t \sim \gff^{\A}(\tilde{\F}_t, s_t)$
    \State Sample $q_t \sim \text{Bernoulli}(k/T)$ 
    \State Take action $a_t$, make query decision $q_t$, and observe $a_t^m q_t$
\EndFor
\end{algorithmic}
\caption{The first reduction: given a full-feedback algorithm and an expected query budget, this algorithm obtains good regret while respecting the query budget.}
\label{alg:active-to-full}
\end{algorithm}

\begin{restatable}[First reduction]{theorem}{thmActive}\label{thm:active-to-full}
Assume that $R: \bbn \times [0,1] \to \bbrpos$ is concave in its first argument and that $\gff$ is a full-feedback algorithm which satisfies $\rsa(T,\gff,\V,\ell) \le R(T,\sigma)$ for all $T\in\bbn, \sigma \in[0,1],$ and $\V \in \scrv{\sigma}$. Then for any $T \in \bbn, k \in (0,T], \sigma \in[0,1],$ and $\V \in \scrv{\sigma}$, $\gk$ satisfies $Q(T,\gk,\V) = k$ and
\[
\rsa(T,\gk,\V,\ell) \le \frac{T}{k} R(k,\sigma).
\]
\end{restatable}
The proof of \Cref{thm:active-to-full} can be found in \Cref{sec:active-to-full-proof}.

\subsection{Applying the Reduction}\label{sec:active-to-full-consequences}

To apply this reduction, we need a full-feedback algorithm. Many such algorithms exist, with typical regret bounds being roughly $O(\sqrt{T})$.\footnote{Theorem 21.10 in \citet{shalev-shwartz_understanding_2014} is the canonical general bound. See \citet{block2022smoothed,haghtalab2022oracle,haghtalab2024smoothed} for the $\sigma$-smooth case.} \Cref{thm:active-to-full} can be applied to these results, but the resulting bounds are not good enough for our final reduction to produce sublinear regret in MDPs. Fortunately, we can obtain better bounds if we assume that $\Pi$ contains a policy with zero loss.  This assumption is called \emph{realizability}. In our case, the loss function will be based on predicting the mentor's actions, so the mentor policy will have zero prediction loss, satisfying realizability. However, we do \emph{not} assume that the mentor is optimal with respect to the MDP reward function that is our ultimate goal.

Lemma~\ref{lem:littlestone} is due to \citet{littlestone1988learning}; see also Lemma 21.7 in \citet{shalev-shwartz_understanding_2014}. Lemma~\ref{lem:realizable-alg} is folklore, but we provide a proof in \Cref{sec:realizable-alg}.\looseness=-1

\begin{lemma}
\label{lem:littlestone}
Assume $\exists \pi^* \in \Pi$ such that $\pi^*(s_t) = a_t^m$ $\forall t \in [T]$. If $\Pi$ has Littlestone dimension $d$, then there exists a full-feedback algorithm $\gff$ such that for any $T \in \bbn$ and $\V \in \scrv{0}$,\looseness=-1
\[
\rsa(T,\gff,\V,\ell_{\bfone}) \le d.
\]
\end{lemma}

Assumptions of finite Littlestone or VC dimension traditionally imply that $|\A|=2$, so for brevity, we do not additionally state the $|\A|=2$ assumption. While there exist multiclass generalizations of these dimensions \citep{daniely2015multiclass, natarajan_learning_1989}, they generally do not handle $\sigma$-smoothness. As such, we will use traditional Littlestone and VC dimension.

\begin{restatable}{restatablelemma}{lemRealizableAlg}
\label{lem:realizable-alg}
Assume $\exists \pi^* \in \Pi$ such that $\pi^*(s_t) = a_t^m$ $\forall t \in [T]$. If $\Pi$ has VC dimension $d$, there exists a full-feedback algorithm $\gff$ such that for any $\sigma \in (0,1]$ and $\V \in \scrv{\sigma}$,
\[
\rsa(T,\gff,\V,\ell_{\bfone}) \in O(d \log T + \sigma^+).
\]
\end{restatable}

Letting $R(T,\sigma) = O(d\log T + \sigma^+)$, we obtain the following corollary of \Cref{thm:active-to-full}. (Note that $d\log T + \sigma^+$ is indeed concave in  $T$.) Corollary \ref{cor:active-to-full} is the primary result we will use to instantiate our later reductions to obtain a complete no-regret guarantee for MDPs.

\begin{corollary}[Main active learning bounds]
\label{cor:active-to-full}
Assume $\exists \pi^* \in \Pi$ such that $\pi^*(s_t) = a_t^m$ $\forall t \in [T]$ and that either (1) $\Pi$ has VC dimension $d$ and $\sigma \in (0,1]$ or (2) $\Pi$ has Littlestone dimension $d$ and $\sigma = 0$. Then for any $k\in (0,T]$, there exists an algorithm $\gk$ such that for any $ \V \in \scrv{\sigma}$ we have $Q(T,\gk, \V) = k$ and
\[
\rsa(T,\gk,\V,\ell_{\bfone}) \in O\left(\frac{T}{k}(d\log k + \sigma^+)\right).
\]
\end{corollary}

Corollary \ref{cor:active-to-full} assumes that $|\A|=2$ (this is implied by finite VC or Littlestone dimension), but this assumption can easily be relaxed via the ``one versus rest'' reduction. Essentially, one runs a separate copy of the algorithm for each $a \in \A$ where each copy's goal is to predict whether $a_t^m = a$. If all copies are correct, then we can fully determine $a_t^m$. Since each copy is given a binary prediction task, Corollary \ref{cor:active-to-full} applies. Thus by the union bound, the total regret is at most the sum of regret bounds across the $|\A|$ copies, and similarly for queries. These bounds can likely be improved by, e.g., applying multiclass variants of VC or Littlestone dimension, but this is beyond the scope of the paper. See Appendix C of \citet{plaut2025avoiding} or Chapter 29 of \citet{shalev-shwartz_understanding_2014} for more details. \looseness=-1


\section{Second Reduction: Avoiding Catastrophe $\to$ Standard Adversarial Active Learning}\label{sec:ac-to-standard}

Our second reduction shows that any algorithm which performs well in the standard setting can be transformed into an algorithm which avoids catastrophe. The basic idea is to follow the original algorithm if the current state is ``familiar'' (defined by a small distance to previous relevant queries) and otherwise ask for help. To formalize this, we must define our model of avoiding catastrophe in online learning.

\subsection{Our Model of Avoiding Catastrophe in Online Learning}\label{sec:model-ac}

In this model, the ``correct'' actions are determined by a mentor policy $\pi^m \in \Pi$ which is chosen upfront by the adversary. Formally, $a_t^m = \pi^m(s_t)$ for all $t \in [T]$. Conceptually, the adversary generates the environment, which includes the mentor policy. Once the mentor policy is chosen, the mentor simply reports $\pi^m(s_t)$ when queried.

    Unlike before, the goal is not simply to predict the mentor's actions but to maximize a sequence of unknown reward functions $\bfmu = (\mu_1,\dots,\mu_T) \in (\s \times \A \to [0,1])^T$. However, these are not ``normal'' reward functions: instead, $\mu_t(s_t,a_t)$ represents the probability of avoiding catastrophe at time $t$ (conditioned on no prior catastrophe). Then the agent's overall chance of avoiding catastrophe is $\prod_{t=1}^T \mu_t(s_t,a_t)$. As before, we encourage the reader to think of ``catastrophe'' as an irreparable error; indeed, our final reduction will choose $\bfmu$ to capture certain transition probabilities. However, the model is valid for any choice of $\bfmu$. Importantly, we do not assume that $\pi^m$ is optimal for $\bfmu$.\footnote{Realizability only states that $\pi^m \in \Pi$; it says nothing about how much reward $\pi^m$ will generate.}

We wish to minimize the following multiplicative regret objective:\looseness=-1
\begin{align*}
\rmul(T, \Gamma, \V, \bfmu) =&\ \E\left[\log \prod_{t=1}^T \mu_t(s_t, \pi^m(s_t)) - \log \prod_{t=1}^T \mu_t(s_t, a_t)\right].
\end{align*}
In other words, the agent should avoid catastrophe nearly as well as the mentor. For brevity, let $\mu_t^m(s) = \mu_t(s,\pi^m(s))$ for $s \in \s$. 



We assume that the agent never directly observes rewards: the only feedback it receives is from queries to the mentor. This is because in the real world, one never observes the true probability of catastrophe: only whether catastrophe occurred. (One may be able to detect “close calls” in some cases, but observing the precise probability seems unrealistic.)

The logarithms in the regret definition are included for consistency with the literature (e.g., Chapter 9 of \citealp{cesa2006prediction}) but do necessitate a special case for rewards of zero. One solution is to assume that rewards cannot be exactly zero. For an adversary $\V$, let $\mum(\V)$ be the infimum of possible rewards across all possible states, actions, and time steps given $\V$. Typically $\V$ will be clear from context and we will just write $\mum$. If $\mum > 0$, then $\rmul$ is always well-defined. We will explicitly note when we assume $\mum > 0$, since this is slightly restrictive and our ultimate MDP results do not require this.\looseness=-1

One can also consider a more standard additive regret objective:
\begin{align*}
\rplus(T, \Gamma, \V, \bfmu) =&\ \E\left[\sum_{t=1}^T \mu_t(s_t, \pi^m(s_t)) - \sum_{t=1}^T \mu_t(s_t, a_t)\right].
\end{align*}
This objective can be interpreted as minimizing the ``total amount of irreversible damage'' rather than the probability of a single catastrophic event.
 
For both objectives, we want the \emph{total} regret to go to 0 as $T\to\infty$, not just the average. This is because we want the total chance of catastrophe to go to 0, not just the average. Thus the goal in this model is \emph{subconstant} regret instead of sublinear regret. This may be counterintuitive, since the regret is a sum of nonnegative quantities and thus is nondecreasing with $t$ (at least for an optimal mentor). However, subconstant regret means that the total regret goes to 0 as a function of the time horizon $T$, not the time step $t$. Choosing algorithmic parameters as a function of $T$ is the key to subconstant regret.


\subsection{Local Generalization} We assume that $\bfmu$ and $\pi^m$ satisfy $L$-\emph{local generalization}. Informally, if the mentor told us that action $a$ is safe in state $s'$, then $a$ is probably also safe in a similar state $s$. Formally, let $\s \subseteq \bbr^n$ (one could also allow a generic metric space). We assume that there exists $L > 0$ such that for all $s,s' \in \s$ and $t \in [T]$, $|\mu_t^m(s) - \mu_t(s, \pi^m(s'))| \le L \norm{s-s'}$, where $||\cdot||$ denotes Euclidean distance.  This represents the ability to transfer knowledge between similar states:
\[
\big|\!\!\underbrace{\mu_t(s, \pi^m(s))}_{\text{Taking the right action}} -\! \underbrace{\mu_t(s, \pi^m(s'))}_{\text{Using what you learned in $s'$}}\!\!\!\!\!\!\!\big| \, \le\: \underbrace{L\norm{s-s'}}_{\text{State similarity}}.
\]
For $X \subseteq \s$, define $\diam(X) = \max_{s, s' \in X} \norm{s-s'}$. For brevity, we write $\diam(\smols) = \diam(\{s_1,\dots,s_T\})$. Let $\calU(L)$ be the set of all $(\bfmu,\pi^m)$ pairs satisfying $L$-local generalization.

The local generalization assumption is vital to our results: this is what allows us to detect when a state is unfamiliar. As such, we now take some time to justify this assumption in detail.\looseness=-1

First, the ability to transfer knowledge between similar states seems fundamental to intelligence and is well-understood in psychology (e.g., \citealp{esser_actioneffect_2023}) and education  (e.g., \citealp{hajian_transfer_2019}). Crucially, the state space $\s\subseteq \bbr^n$ can be any encoding of the agent’s situation, not just its physical positioning. For example, a 3 mm spot and a 3.1 mm spot on X-rays likely have similar risk levels for cancer (assuming similar density, location, etc.). If the risk level abruptly increases for any spot over 3 mm, then local generalization may not hold for a naive encoding which treats size as a single dimension. However, a more nuanced encoding would recognize that these two situations---a 3 mm vs 3.1 mm spot---are in fact \emph{not} similar. \looseness=-1

Constructing a suitable encoding may be challenging, but we do \emph{not} require the agent to have explicit access to such an encoding, nor does it need to know $L$: it only needs a nearest-neighbor distance oracle. Formally, the agent needs to compute $\min_{s \in X}\norm{s_t-s}$, where $X\subseteq \s$ is a particular subset of previously queried states. Informally, the agent only needs the ability to detect unfamiliar states. While this task remains far from trivial, we argue that it is more tractable than fully constructing a suitable encoding. See \Cref{sec:conclusion} for a discussion of potential future work on this topic.\looseness=-1

We note that these encoding-related questions apply similarly to the more standard assumption of Lipschitz continuity. In fact, Lipschitz continuity implies local generalization when the mentor is optimal (Proposition E.1 in \citealp{plaut2025avoiding}). We also mention that without local generalization, avoiding catastrophe is impossible even when the mentor policy class has finite VC dimension and the adversary is $\sigma$-smooth (Theorem E.2 in \citealp{plaut2025avoiding}).\looseness=-1


\subsection{The Reduction}

We can now define our reduction. Given access to a ``default'' algorithm $\Gamma$ which performs well in the absence of catastrophe, \Cref{alg:ac-to-standard} first computes what $\Gamma$ would do in the current state (based on a simulated history of what $\Gamma$ would have done in previous states). If the current state is ``familiar'' (i.e., has a small nearest-neighbor distance $\min_{(s,a) \in X: a = \tilde{a}_t} \norm{s_t - s}$), we follow $\Gamma$. Otherwise, we ask for help. The ``familiarity'' threshold $\ep$ must be carefully chosen to balance safety with self-sufficiency.  Ultimately we choose $\ep = T^\frac{-1}{n+1}$ (recall that $n$ is the dimension of $\s$). However, we first prove bounds for general $\ep$. Similar to the first reduction, Algorithm~\ref{alg:ac-to-standard} requires knowledge of $T$, but this can be removed with standard doubling tricks.
\looseness=-1

\begin{algorithm}[h]
\begin{algorithmic}[1]
\State Inputs: Algorithm $\Gamma,\,  \ep \in \bbrspos,\, T \in \bbn$
\State $X \gets \emptyset$
\For{$t$ \textbf{from} $1$ \textbf{to} $T$}
\State Observe $s_t$ from adversary
\State $\tilde{\F}_t \gets \big(s_i,\tilde{a}_i,\pi^m(s_i)\tilde{q}_i\big)_{i \in [t-1]}$ \Comment{Simulated history for $\Gamma$}
\State Sample $\tilde{q}_t \sim \Gamma^Q(\tilde{\F}_t, s_t)$
\State Sample $\tilde{a}_t \sim \Gamma^{\A}(\tilde{\F}_t,s_t, 0)$
    \If {$\min_{(s,a) \in X: a = \tilde{a}_t} \norm{s_t - s} > \ep$} \Comment{Out-of-distribution: ask for help}
                        \State $q_t \gets 1$, observe $\pi^m(s_t)$ 
                        \State $a_t \gets \pi^m(s_t)$
                        \State $X \gets X \cup\{(s_t, \pi^m(s_t))\}$   
                \Else  \Comment{In-distribution: follow $\Gamma$}
                            \State $q_t \gets \tilde{q}_t$, observe $\pi^m(s_t) q_t$
                            \State $a_t \gets \tilde{a}_t$
\EndIf
\EndFor
\end{algorithmic}
\caption{The second reduction: given an algorithm which performs well in the absence of catastrophe, this algorithm avoids catastrophe using a limited number of additional queries.\looseness=-1}
\label{alg:ac-to-standard}
\end{algorithm}

\newpage 
\begin{restatable}[Second reduction]{theorem}{thmACToStandard}\label{thm:ac-to-standard}
Let $k \in \bbrpos$ and $R:\bbrpos\times\bbn\times [0,1]\to\bbrpos$. Let $\Gamma$ be a query-agnostic algorithm satisfying $Q(T,\Gamma,\V) \le k$ and $\rsa(T,\Gamma,\V,\ell_{\bfone}) \le R(k,T,\sigma)$ for all $T\ge k,\sigma \in [0,1],$ and $\V \in \scrv{\sigma}$. 
Let $\gac$ denote  \Cref{alg:ac-to-standard} with inputs $\Gamma,$ $T \ge k$, and $\ep > 0$. Then for any $\sigma \in [0,1], \V \in \scrv{\sigma}$, and $(\bfmu,\pi^m) \in \calU(L)$,
\begin{align*}
\rmul(T, \gac, \V, \bfmu) \le&\ \frac{L\ep R(k,T,\sigma) }{\mum} \quad \textnormal{if } \mum > 0, \\
\rplus(T, \gac, \V, \bfmu) \le&\ L \ep R(k,T,\sigma),\\
Q(T,\gac,\V) \in&\ O\left(k +R(k,T,\sigma)  + \frac{|\A| \E[\diam(\smols)^n]}{\ep^n}\right).
\end{align*}
\end{restatable}


The proof of \Cref{thm:ac-to-standard} appears in \Cref{sec:ac-to-standard-proof}. To obtain subconstant regret and sublinear queries, we apply \Cref{thm:ac-to-standard} to Corollary~\ref{cor:active-to-full} with $k=T^\frac{2n+1}{2n+2}$ and $\ep = T^\frac{-1}{n+1}$. For readability, we have simplified the bounds at the cost of slightly looser bounds.

\begin{restatable}[Final avoiding catastrophe bounds]{theorem}{thmACToStandardFinal}\label{thm:ac-to-standard-final}
Assume that either (1) $\Pi$ has VC dimension $d$ and $\sigma \in (0,1]$ or (2) $\Pi$ has Littlestone dimension $d$ and $\sigma = 0$. Let $k =T^\frac{2n+1}{2n+2}$ and let $\Gamma_k$ be the corresponding algorithm from Corollary~\ref{cor:active-to-full}. If $\gac$ denotes Algorithm~\ref{alg:ac-to-standard} with inputs $\Gamma_k, T,$ and $\ep = T^\frac{-1}{n+1}$, then
\begin{align*}
\rmul(T, \gac, \V, \bfmu) \in&\ O\left(\frac{L}{\mum} T^\frac{-1}{2n+2} (d\log T + \sigma^+)\right)\quad \textnormal{if }\mum > 0, \\
\rplus(T, \gac, \V, \bfmu) \in&\ O\left(L T^\frac{-1}{2n+2} (d\log T + \sigma^+)\right), \\
Q(T,\gac,\V) \in&\ O\left(  T^\frac{2n+1}{2n+2}\left(d+ \sigma^+ +  \E[\diam(\smols)^n]\right)\right),
\end{align*}
for any $(\bfmu,\pi^m) \in \calU(L)$ and $\V \in \scrv{\sigma}$.
\end{restatable}

\begin{remark}\label{rem:simultaneous}
\textnormal{Because the agent never observes rewards, the distribution of $(\smols,\smola)$ depends only on the adversary $\V$ (which determines $\pi^m$) and the algorithm $\gac$, and not on $\bfmu$. Hence this \emph{single} distribution of $(\smols,\smola)$ satisfies the bounds in \Cref{thm:ac-to-standard-final} \emph{simultaneously} for all admissible choices of $\bfmu$. In other words, run the online learning protocol for $T$ time steps without choosing $\bfmu$. Then, at the end, evaluate $\sup_{\bfmu:(\bfmu,\pi^m)\in\calU(L)} \rmul(T,\gac,\V,\bfmu)$, and similar for $\rplus$. This property will be important for the third reduction.}
\end{remark}

\section{Third Reduction: Sublinear Regret in MDPs $\to$ Avoiding Catastrophe}\label{sec:mdp-to-ac}

Our third reduction shows that any algorithm which avoids catastrophe (i.e., has subconstant regret in the model from \Cref{sec:model-ac}) guarantees sublinear regret in multichain MDPs. This leads to our ultimate goal: a no-regret guarantee for multichain MDPs using sublinear mentor queries. \looseness=-1

\subsection{Online Learning in MDPs with a Mentor}\label{sec:model-mdp}

MDPs involve a type of constrained adversary who only chooses a transition function $P: \s\times\A\to\Delta(\s)$, a mentor policy $\pi^m: \s \to \A$, and an initial state distribution $\D_1 \in \Delta(\s)$. Then $s_1\sim \D_1$ and for all $t > 1$, $s_t$ is sampled from $P(s_{t-1}, a_{t-1})$. As in \Cref{sec:model-ac}, $a_t^m = \pi^m(s_t)$ for all $t \in [T]$. Sometimes it will be useful to denote the adversary as $\V = (P, \D_1,\pi^m)$, in which case we can also write $(P,\D_1,\pi^m) \in \scrv{\sigma}$.

The agent's objective is to maximize a reward function $r:\s\times\A\to[0,1]$. As in \Cref{sec:model-ac}, we do not assume that the mentor is optimal. Also like \Cref{sec:model-ac}, the agent here never directly observes rewards and only receives feedback from queries.\footnote{In the typical MDP framework, the agent does observe rewards directly. However, our algorithms do not use these observations, so we might as well strengthen our results by removing such observations.}\looseness=-1

Unlike the other models, the transition function allows us to define a distribution of \emph{mentor states} $\sm = (s_1^m,\dots,s_T^m)$, where $s_1^m \sim \D_1$ and for $t > 1$, $s_t^m \sim P(s_{t-1}^m, \pi^m(s_{t-1}^m))$. Thus we can define MDP regret by comparing the expected reward of the agent and mentor on their \emph{respective} sequences of states:
\[
\rmdp(T,\Gamma,(P,\D_1,\pi^m), r) = \E\left[\sum_{t=1}^T r(s_t^m, \pi^m(s_t^m) )- \sum_{t=1}^T r(s_t, a_t)\right].
\]
This corresponds to independently running the mentor and the agent each for $T$ steps from the same initial state and comparing their expected total reward. For example, recall \Cref{fig:heaven_hell_problem} and suppose the mentor goes to Heaven but the agent goes to Hell. Our definition of regret appropriately penalizes such an agent: for all $t > 1$, we have $s_t^m = \text{Heaven}$ while $s_t = \text{Hell}$, resulting in a very poor regret of $T-1$.

We assume that $P$ and $\pi^m$ satisfy $L$-local generalization with respect to total variation distance: $||P(s,\pi^m(s)) - P(s, \pi^m(s'))||_{TV} \le L\norm{s-s'}$ for any $s,s' \in \s$. (We assume $\s \subseteq \bbr^n$ and define $\diam(X)$ as in \Cref{sec:model-ac}.) The most general version of the third reduction will also require $L$-local generalization for $r$:  $|r(s,\pi^m(s)) - r(s,\pi^m(s'))| \le L\norm{s-s'}$ for any $s,s' \in \s$. We assume that $L$ is the same for both $P$ and $r$ (if not, simply use the maximum). Let $\p(L)$ and $\R(L)$ denote the sets of $(P,\pi^m)$ and $(r,\pi^m)$ pairs satisfying $L$-local generalization, respectively.

\subsection{The Reduction}

This reduction requires no modification of the algorithm: any algorithm which has subconstant regret for $\rplus$ must have certain properties which ensure that it will also perform well in multichain MDPs.\footnote{We reduce to $\rplus$ instead of $\rmul$ to avoid dealing with $\mum$.} Specifically, such an algorithm will have MDP regret at most $(T+1) \cdot \rplus$, as stated by \Cref{thm:mdp-to-ac}. If $\rplus$ is subconstant, then the resulting MDP regret is sublinear.

\begin{restatable}[Third reduction]{theorem}{thmMDPToAC}
\label{thm:mdp-to-ac}
Fix some $R:\bbn\times[0,1]\to\bbrpos$ and let $\Gamma$ be an algorithm which satisfies $\rplus(T,\Gamma,\V,\bfmu) \le R(T,\sigma)$ for all $T \in \bbn, \sigma \in [0,1], \V \in \scrv{\sigma},$ and $ (\bfmu, \pi^m) \in \calU(L)$. Then $\Gamma$ satisfies
\[
\rmdp(T,\Gamma,(P,\D_1,\pi^m), r) \le (T+1) R(T,\sigma)
\]
for all $T \in \bbn, \sigma \in [0,1],$ and $ (P,\D_1,\pi^m) \in \scrv{\sigma}$ with $(P,\pi^m) \in \p(L)$ and $(r,\pi^m) \in \R(L)$.
\end{restatable}

If we can also reduce to the standard adversarial setting---that is, we have access to a bound on $\rsa$---then we do not need local generalization for $r$.

\begin{restatable}[Third reduction: no local generalization for $r$]{theorem}{thmMDPToACNoR}
\label{thm:mdp-to-ac-no-r}
For functions $R_1,R_2:\bbn\times[0,1]\to\bbrpos$, let $\Gamma$ be an algorithm which satisfies $\rsa(T,\Gamma,\V,\ell_{\bfone}) \le R_1(T,\sigma)$ and $\rplus(T,\Gamma,\V,\bfmu) \le R_2(T,\sigma)$ for any $T \in \bbn, \sigma \in [0,1], \V \in \scrv{\sigma},$ and $(\bfmu,\pi^m) \in \calU(L)$.
Then $\Gamma$ satisfies
\[
\rmdp(T,\Gamma,(P,\D_1,\pi^m), r) \le R_1(T,\sigma) + T R_2(T,\sigma)
\]
for any $T \in \bbn, \sigma \in [0,1],$ and $ (P,\D_1,\pi^m) \in \scrv{\sigma}$ with $(P,\pi^m) \in \p(L)$.
\end{restatable}

\Cref{thm:mdp-to-ac-no-r} requires a single algorithm with bounds on both $\rsa$ and $\rplus$. Although $\rsa$ and $\rplus$ originate from different models, one advantage of our unifying online learning framework is that these metrics are well-defined for any algorithm in our framework. Furthermore, in the course of proving \Cref{thm:ac-to-standard}, we actually already proved that $\gac$ (Algorithm~\ref{alg:ac-to-standard}) has bounds on both $\rsa$ and $\rplus$. Specifically, this is stated by Lemma~\ref{lem:ac-num-errors-final}, which we can directly reuse.

The culmination of our work is \Cref{thm:mdp-to-ac-final} below, which establishes the existence of a no-regret algorithm for multichain MDPs with sublinear queries. We use the exact same algorithm as \Cref{thm:ac-to-standard-final}. The bound on $Q(T,\Gamma,\V) = Q(T,\Gamma, (P,\D_1,\pi^m))$ is automatically inherited and the regret bound only requires a short proof. Essentially, we apply \Cref{thm:mdp-to-ac-no-r} with $R_1(T,\sigma)$ as the bound on $\rsa$ from Corollary~\ref{cor:active-to-full} and $R_2(T,\sigma)$ as the bound on $\rplus$ from \Cref{thm:ac-to-standard-final}.\looseness=-1

\begin{restatable}[Final no-regret guarantee for multichain MDPs]{theorem}{thmMDPToACFinal}
\label{thm:mdp-to-ac-final}
Assume that either (1) $\Pi$ has VC dimension $d$ and $\sigma \in (0,1]$ or (2) $\Pi$ has Littlestone dimension $d$ and $\sigma = 0$. Let $k =T^\frac{2n+1}{2n+2}$, let $\Gamma_k$ be the corresponding algorithm from Corollary~\ref{cor:active-to-full}, and let $\gac$ denote Algorithm~\ref{alg:ac-to-standard} with inputs $\Gamma_k, T,$ and $\ep = T^\frac{-1}{n+1}$. Then $\gac$ satisfies
\begin{align*}
\rmdp(T,\gac,(P,\D_1,\pi^m), r) \in&\ O\left(L T^\frac{2n+1}{2n+2} (d\log T + \sigma^+)\right), \\
Q(T,\gac, (P,\D_1,\pi^m)) \in&\ O\left(  T^\frac{2n+1}{2n+2}\left(d +\sigma^+ +  \E[\diam(\smols)^n]\right)\right),
\end{align*}
 for any $(P,\D_1,\pi^m)\in \scrv{\sigma}$ with $(P,\pi^m) \in \p(L)$.
\end{restatable}

For concreteness, we present a consolidated complete algorithm for the $\sigma$-smooth case below. Since our reduction is quite general, this is only one possible algorithm, but it is sufficient for illustrative purposes. For the full-feedback algorithm $\gff$, we will use the algorithm from Lemma~\ref{lem:realizable-alg}. This algorithm consists of forming a $\frac{1}{T}$-cover (see Appendix~\ref{sec:realizable-alg} for the definition) and running the ``exponentially weighted average forecaster'' from Corollary 2.3 of \citet{cesa2006prediction} on this cover. The algorithm assumes that $A = \{0,1\}$.

\begin{algorithm}
\begin{algorithmic}[1]
\State Inputs: {$T \in \bbn,\, k \in (0,T],\, \ep \in \bbrspos$}
\State $\tilpi \gets$ any $\frac{1}{T}$-cover of $\Pi$
\State $X \gets \emptyset$
\For{$\pi \in \tilpi$}
    \State $w_\pi \gets 1/|\tilpi|$
\EndFor
\For{$t$ \textbf{from} $1$ \textbf{to} $T$}
\State Observe $s_t$
\State $p_t \gets \sum_{\pi \in \tilpi} w_\pi \pi(s_t)$
\State Sample $\tilde{a}_t \sim \text{Bernoulli}(p_t)$
\State \texttt{is\_ood} $\gets \big(\min_{(s,a) \in X: a = \tilde{a}_t} \norm{s_t - s} > \ep\big)$
\If {\texttt{is\_ood}} \Comment{Out-of-distribution: ask for help}
    \State $q_t \gets 1$, observe $\pi^m(s_t)$
     \State $a_t  \gets \pi^m(s_t)$
    \State $X \gets X \cup\{(s_t, \pi^m(s_t))\}$
\Else \Comment{In-distribution: follow exponentially-weighted average}
    \State $a_t \gets \tilde{a}_t$
\EndIf
\State Sample $\tilde{q}_t \sim \text{Bernoulli}(k/T)$ \Comment{Update weights w.p. $k/T$, even on OOD rounds}
\If{$\tilde{q}_t = 1$}
\State $q_t \gets 1$, observe $\pi^m(s_t)$
\State $W \gets \sum_{\pi \in \tilpi} w_\pi \exp(-\bfone(\pi^m(s_t) \ne \pi(s_t)))$
    \For{$\pi \in \tilpi$}
    \State $w_\pi \gets \cfrac{w_\pi \exp(-\bfone(\pi^m(s_t) \ne \pi(s_t)))}{W} $
    \EndFor
\EndIf
\If{$\tilde{q}_t = 0$ and not \texttt{is\_ood}}
    \State $q_t \gets 0$
 \EndIf
\EndFor
\end{algorithmic}
\caption{A final concrete algorithm obtained by applying our reduction to the exponentially-weighted average predictor from \citet{cesa2006prediction}.}
\label{alg:final}
\end{algorithm}

\begin{corollary}
Assume that $\Pi$ has VC dimension $d$ and that $\sigma \in (0,1]$. Let $\Gamma_{\textnormal{final}}$ denote Algorithm~\ref{alg:final} with inputs $k = T^{\frac{2n+1}{2n+2}}$ and $\ep = T^{\frac{-1}{n+1}}$. Then $\Gamma_{\textnormal{final}}$ satisfies
\begin{align*}
\rmdp(T,\Gamma_{\textnormal{final}},(P,\D_1,\pi^m), r) \in&\ O\left(L T^\frac{2n+1}{2n+2} (d\log T + \sigma^{-1})\right), \\
Q(T,\Gamma_{\textnormal{final}}, (P,\D_1,\pi^m)) \in&\ O\left(  T^\frac{2n+1}{2n+2}\left(d +\sigma^{-1} +  \E[\diam(\smols)^n]\right)\right),
\end{align*}
 for any $(P,\D_1,\pi^m)\in \scrv{\sigma}$ with $(P,\pi^m) \in \p(L)$.
\end{corollary}

\subsection{Key Proof Idea}\label{sec:mdp-proof-idea}

The key idea is to decompose regret into \emph{state-based regret} and \emph{action-based regret} by simply adding and subtracting $\E[\sum_{t=1}^T r(s_t, \pi^m(s_t))]$:
\begin{align*}
\E\left[\sum_{t=1}^T r(s_t^m, \pi^m(s_t^m)) - \sum_{t=1}^T r(s_t, a_t)\right] =
\end{align*}
\begin{align*}
\underbrace{\E\left[\sum_{t=1}^T r(s_t^m, \pi^m(s_t^m)) - \sum_{t=1}^T r(s_t, \pi^m(s_t))\right]}_{\textnormal{State-based regret}} + \underbrace{\E\left[\sum_{t=1}^T r(s_t, \pi^m(s_t)) - \sum_{t=1}^T r(s_t, a_t)\right]}_{\textnormal{Action-based regret}}. 
\end{align*}

\emph{State-based regret} measures how bad the agent's states $\smols$ are compared to the mentor's states $\sm$. Here ``bad'' is evaluated based on the actions the mentor would take in each state. To bound the state-based regret, we use the local generalization of $P$ to bound the deviation between the distributions of $\smols$ and $\sm$ as measured by $\sup_{X\subseteq \s}(\Pr[s_t^m \in X] - \Pr[s_t \in X])$. Most of the proof is focused on showing that the state-based regret is at most $T \rplus$.\looseness=-1

\emph{Action-based regret} measures how bad the agent's actions are compared to the mentor's, evaluated on the agent's states $\smols$. Local generalization of $r$ will imply that the action-based regret is at most $\rplus$. Combining this with our bound on state-based regret yields  $\rmdp \le (T+1)\rplus$, as desired.

To our knowledge, this decomposition is novel and could be of independent interest. Superficially, our regret decomposition might resemble existing techniques like the reference-advantage decomposition \citep{zhang2020almost}. The key difference is that each term in our decomposition includes only short-term rewards, and long-term effects remain implicit. In contrast, prior decompositions (including reference-advantage) typically use $Q$-functions and/or value functions which explicitly capture the long-term value of policies/states/actions. Neither approach is necessarily ``better'', but they may be suitable for different contexts.\looseness=-1

\section{Conclusion}\label{sec:conclusion}

In this paper, we provide a sequence of three reductions which produce (to our knowledge) the first no-regret guarantee for multichain MDPs. Conceptually, we prove that this algorithm obtains high reward while becoming self-sufficient, even when errors may be catastrophic.\looseness=-1

Although we think these insights are broadly applicable, our algorithm is not ready for practical usage. For one, computing the $\frac{1}{T}$-cover is computationally expensive. There exist techniques for avoiding computing such covers explicitly (see, e.g., \citealp{block2022smoothed}), and this could be a topic for future work.

Also, our final query and regret bounds of $\tilde{O}(T^\frac{2n+1}{2n+2})$ are only barely sublinear in $T$ when $n$ is large. The exponent arises from the number of queries needed to cover an $n$-dimensional space; essentially, the curse of dimensionality. Another factor is that our agent only learns from querying. Providing an initial offline data set would remove the need to cover the entire space with mentor queries and could significantly improve our bounds. \looseness=-1

Another limitation is that the algorithm relies on perfectly computing distances between states in order to exploit local generalization. This is analogous to requiring a perfect out-of-distribution (OOD) detector, which remains a major open problem \citep{yang2024generalized}. Future work might consider a more realistic model of OOD detection and/or a less strict version of the local generalization assumption. Our work also makes the standard yet limiting assumptions of total observability and knowledge of the policy class, which could be relaxed in future work.

We are also interested in cautious learning without a mentor, since a mentor may not always be available. One possibility is learning from bad-but-not-catastrophic experiences: for example, someone who gets serious (but not fatal) food poisoning may be more cautious with food safety in the future. This phenomenon is not captured in our current model.\looseness=-1


More broadly, we think that the principle of acting cautiously when uncertain may be more powerful than previously suspected. We are hopeful that this idea can help make AI systems safer and more beneficial for all of society.




\section*{Acknowledgments and Disclosure of Funding}

This work was supported by a gift from Open Philanthropy to the Center for Human-Compatible AI at UC Berkeley. We have no competing interests to declare. 

We would like to thank (in alphabetical order) Aly Lidayan, Bhaskar Mishra, Cameron Allen, Daniel Jarne Ornia, Karim Abdel Sadek, Matteo Russo, Michael Cohen, Nika Haghtalab, Ondrej Bajgar, Scott Emmons, and Tianyi Alex Qiu for feedback and discussion which significantly improved the paper. \looseness=-1


\appendix

\section{First Reduction Proofs}\label{sec:active-to-full-proof}

This section presents the proofs for our first reduction (\Cref{thm:active-to-full}).

\subsection{Proof Notation}
In the proof, we fix an arbitrary $T \in \bbn, \sigma \in [0,1], \V \in \scrv{\sigma}$ and use the following notation:
\begin{enumerate}[leftmargin=1.5em,
    topsep=0.5ex,
    partopsep=0pt,
    parsep=0pt,
    itemsep=0.5ex]
    \item Let $K = \sum_{t=1}^T q_t$  denote the realized number of queries. Note that $\E[K] = k$.
    \item Let $t_1 < t_2 < \dots < t_K$ be the time steps $t \in [T]$ where $q_t = 1$.
    \item   For each $t \in [T]$, define $\till_t(a,a^m) = q_t \frac{T}{k} \ell(a,a^m)$.
    \item Let $\Delta_T(\gk,\pi) =  \sum_{t=1}^T \ell(a_t, a^m_t) -\sum_{t=1}^T \ell(\pi(s_t), a^m_t)$ for each $\pi \in \Pi$. Note that\\ $\rsa(T,\gk,\V,\ell) = \sup_{\pi \in \Pi}\, \E[\Delta_T(\gk,\pi)]$.
    \item Let $\tilD_T(\gk,\pi) = \sum_{t=1}^T \till_t(a_t,a^m_t) - \sum_{t=1}^T \till_t(\pi(s_t), a^m_t)$. 
\end{enumerate}
The trickiest part of the proof is proving the equivalence between running $\gk$ on the true instance $\I = (T,\V, \Pi, \ell)$ and running $\gff$ on a different instance $\I^{\U}$, where $\U \in\{0,1\}^T$ is a possible realization of $\q$. Formally, for each $\sigma \in [0,1]$, $\V \in \scrv{\sigma},$ and $\U \in \{0,1\}^T$, define $\I^{\U} = (T^{\U},\V^{\U},\Pi, \ell, \gff)$ as follows. 
\begin{enumerate}[leftmargin=1.5em,
    topsep=0.5ex,
    partopsep=0pt,
    parsep=0pt,
    itemsep=0.5ex]
    \item Let $T^{\U} = \sum_{t=1}^T u_t$.
    \item For a history $\F$ of length $j-1$, define $\V^{\U}(\F)= \D(s_{t_j}, a^m_{t_j} \mid  \F_{t_j} \cap \U = \F,\, \q = \U)$ if $\Pr[\F_{t_j} \cap \U = \F,\, \q = \U] > 0$. Otherwise, let $\V^{\U}(\F)$ be any distribution whose marginal distribution on $\s$ is $\sigma$-smooth.
    \item For clarity, let $s_t,a^m_t,a_t, q_t,$ and $\F_t$ respectively denote the state, correct action, algorithm's action, query decision, and history at time $t$ in $\I$, i.e., as induced by $\V$ and $\gk$. Let $s_t^{\U},a_t^{m,\U},a_t^{\U},q_t^{\U},$ and $\F_t^{\U}$ denote the analogous random variables in $\I^{\U}$, i.e., as induced by $\V^{\U}$ and $\gff$.\footnote{Since $\gff$ is a full-feedback algorithm, $q_t^{\U}$ is always 1, but we include it for completeness.}  
\end{enumerate}

\subsection{Proof}

Lemma \ref{lem:unbiased} shows that $\tilD_T$ is an unbiased estimator of $\Delta_T$. Lemma \ref{lem:smoothness} ensures that $\V^{\U}$ is $\sigma$-smooth, which is necessary in order to apply the assumed regret bound for $\gff$. Lemma \ref{lem:standard-equiv} is the key to the proof: it states that the distribution of histories is the same for $\I$ and $\I^{\U}$ (under a particular transformation). To complete the proof of \Cref{thm:active-to-full}, we apply the regret bound for $\gff$ to $\I^{\U}$, which we can then lift to a regret bound for $\gk$ on the original instance $\I$ using Lemmas \ref{lem:unbiased} and \ref{lem:standard-equiv}.

\begin{lemma}\label{lem:unbiased}
For all $\pi\in\Pi$, $\E[\tilD_T(\gk, \pi)] = \E[\Delta_T(\gk,\pi)]$.
\end{lemma}
\begin{proof}
Fix a $t \in [T]$ and let $a$ be any random variable taking values in $\A$ such that $q_t$ is independent of the pair $(a, a^m_t)$. Then $q_t$ is also independent of $\till_t(a,a_t^m)$ so 
\begin{align*}
\E[\till_t(a, a^m_t)] =& \E[q_t] \E\left[\frac{T}{k}\ell(a, a^m_t)\right]
= \frac{k}{T} \frac{T}{k} \E[\ell(a, a^m_t)]
= \E[\ell(a, a^m_t)].
\end{align*}
Since $q_t \sim \text{Bernoulli}(k/T)$, we know that $q_t$ is independent of both $(a_t, a^m_t)$ and $(\pi(s_t), a^m_t)$. Therefore we can apply $\E[\till_t(a, a^m_t)] = \E[\ell(a, a^m_t)]$ with both $a = a_t$ and $a = \pi(s_t)$, so
\begin{align*}
\E[\Delta_T(\gk,\pi)] =&\ \sum_{t=1}^T \E[\ell(a_t, a^m_t)]-\sum_{t=1}^T \E[\ell(\pi(s_t), a^m_t)]\\
=&\ \sum_{t=1}^T \E[\till_t(a_t, a^m_t)]-\sum_{t=1}^T \E[\till_t(\pi(s_t), a^m_t)] \\
=&\ \E[\tilD_T(\gk,\pi)]
\end{align*}
as required.
\end{proof}

\begin{lemma}[Tower property]
\label{lem:tower}
For random variables $X,Y$ and an event $E$, if $\E_X[X\mid E]$ and $\E_Y[\E_X[X\mid Y,E]\mid E]$ both exist, then $\E_X[X\mid E]=\E_Y[\E_X[X\mid Y,E]\mid E]$.
\end{lemma}

Although the subscripts of the expectations in Lemma~\ref{lem:tower} are not technically necessary, we include them to improve readability.

\begin{lemma}
    \label{lem:smoothness}
For each $\U \in \{0,1\}^T$, the adversary $\V^{\U}$ is $\sigma$-smooth.
\end{lemma}

\begin{proof}
Fix some $j \in [T^{\U}]$,  $\F \in \scrf$, and $X\subseteq \s$. If $\sigma = 0$, the claim is trivial. If $\Pr[\F_{t_j} \cap \U = \F,\, \q = \U] = 0$, then the claim holds by definition of $\V^{\U}$. Thus assume $\sigma > 0$ and $\V^{\U}(\F) = \D(s_{t_j}, a^m_{t_j} \mid  \F_{t_j} \cap \U = \F,\, \q = \U)$. Let $E = \{\F_{t_j} \cap \U = \F,\, \q = \U\}$. Applying Lemma~\ref{lem:tower} with $X = \bfone(s_{t_j} \in X)$ and $Y = \F_{t_j}$ gives us
\begin{align*}
\Pr_{(s,a) \sim \V^{\U}(\F)} [s \in X] = \Pr[s_{t_j} \in X \mid  \F_{t_j} \cap \U = \F,\, \q = \U] 
= \E_{\F_{t_j}}\big[\Pr[s_{t_j} \in X \mid  E, \F_{t_j}]\ \big|\ E\big].
\end{align*}
We claim that $\Pr[s_{t_j} \in X \mid  E, \F_{t_j}] = \Pr[s_{t_j} \in X\mid \F_{t_j}]$. Define events $E_1=\{\F_{t_j} \cap \U=\F\}, E_2 = \{q_i=u_i\ \forall i < t_j\},$ and $E_3 = \{q_i = u_i\ \forall i \ge t_j\}$. Notice that all probability information in $E_1$ and $E_2$ is already contained in $\F_{t_j}$, and that $\F_{t_j}$ is independent of $E_3$.\footnote{Technically $E_1$ depends on $\U$, and $\U$ is not determined by $\F_{t_j}$. However, $\U$ is a constant in our setup.} Thus\looseness=-1
\begin{align*}
\E_{\F_{t_j}}\big[\Pr[s_{t_j} \in X \mid  E, \F_{t_j}] \mid E\big] =&\ \E_{\F_{t_j}}\big[\Pr[s_{t_j} \in X \mid  E_3, \F_{t_j}] \mid E\big] && (\text{$E_1,E_2$ contained within $\F_{t_j}$})\\
=&\ \E_{\F_{t_j}}\big[\Pr[s_{t_j} \in X \mid  \F_{t_j}] \mid E\big] && (\text{Independence of $E_3, \F_{t_j}$})\\
=&\ \E_{\F_{t_j}}\left[\Pr_{(s,a)\sim \V(\F_{t_j})} [s \in X]\ \Big|\ E\right] && (\text{Definition of $s_{t_j} \sim \V(\F_{t_j})$})\\
\le&\ \E_{\F_{t_j}}\left[\frac{1}{\sigma}\beta(X)\ \Big|\ E\right] && (\text{$\V$ is $\sigma$-smooth})\\
=&\ \frac{1}{\sigma}\beta(X). && (\text{Expectation of a constant})
\end{align*}
Therefore $\Pr_{(s,a) \sim \V^{\U}(\F)} [s \in X] \le \frac{1}{\sigma}\beta(X)$, as required.
\end{proof}

\begin{lemma}
\label{lem:standard-equiv}
For any $g:\s\times\A\times\A \to \bbr$, $\ \E[g(s_j^{\U}, a_j^{m,\U}, a_j^{\U})] = \E[g(s_{t_j}, a^m_{t_j}, a_{t_j}) \mid \q = \U]$.
\end{lemma}

\begin{proof}
We prove by induction that for any $\F \in \scrf$ and $j \in [T^{\U}+1]$, $\Pr[\F_j^{\U} = \F] = \Pr[\F_{t_j} \cap \U = \F \mid \q = \U]$. We trivially have $\F_1^{\U} = \emptyset = \F_{t_1} \cap \U$ always, which satisfies the base case. Now assume that the claim holds for some $j < T^{\U} + 1$. Fix some $\F \in \scrf$ of length $j+1$ where all query decisions are 1. Then we can write $\F = \F' \smallfrown (s,a,a^m)$ for some $\F' \in \scrf$ of length $j$ and $s\in\s,a\in\A,a^m\in\A$. We proceed by case analysis.

\textbf{Case 1:} $\Pr[\F_{t_j} \cap \U = \F',\, \q = \U] = 0$. Then
\begin{align*}
\Pr[\F_{t_j} \cap \U = \F',\, \q = \U] =&\ \Pr[\F_{t_j} \cap \U = \F' \mid \q = \U] \Pr[\q = \U] && (\text{Chain rule of probability})\\
=&\ \Pr[\F_j^{\U} = \F'] \Pr[\q = \U]. && (\text{Inductive hypothesis})
\end{align*}
Since $\Pr[\q = \U] > 0$ for all $\U \in \{0,1\}^T$ and $\Pr[\F_{t_j} \cap \U = \F',\, \q = \U] = 0$, we must have $\Pr[\F_j^{\U} = \F'] = 0 = \Pr[\F_{t_j} \cap \U = \F' \mid \q = \U]$. Since $\F = \F' \smallfrown (s,a,a^m)$, this implies that $\Pr[\F_{j+1}^{\U} = \F] = 0 = \Pr[\F_{t_{j+1}}\cap \q = \F \mid \q=\U]$, which satisfies the induction.

\textbf{Case 2:} $\Pr[\F_{t_j} \cap \U = \F',\, \q = \U] > 0$. Using definitions and the chain rule of probability, 
\begin{align}
&\ \Pr[\F_{j+1}^{\U} = \F]\nonumber\\
=&\ \Pr\big[\F_j^{\U}\smallfrown(s_j^{\U}, a_j^{\U},a_j^{m,\U})= \F\big] \nonumber \\
=&\ \Pr[\F_j^{\U} = \F',\, s_j^{\U} = s,\, a_j^{\U} =a,\, a_j^{m,\U}=a^m]\nonumber\\
=&\ \Pr[a_j^{\U} =a\mid \F_j^{\U}=\F', s_j^{\U}=s, a_j^{m,\U}=a^m]\ \Pr[s_j^{\U}=s, a_j^{m,\U}=a^m \mid \F_j^{\U} =\F']\ \Pr[\F_j^{\U} =\F'].\nonumber
\end{align}
Since $\Pr[\F_{t_j} \cap \U = \F',\, \q = \U] > 0$, the second term $\Pr[s_j^{\U}=s, a_j^{m,\U}=a^m \mid \F_j^{\U} =\F']$  is exactly $\V^{\U}(\F')(s,a^m)$. The first term $\Pr[a_j^{\U} =a\mid \F_j^{\U}=\F', s_j^{\U}=s, a_j^{m,\U}=a^m]$ does not quite match the definition of $\gff^{\A}(\F',s)(a)= \Pr[a_j^{\U} =a\mid \F_j^{\U}=\F', s_j^{\U}=s]$. However, since $\gff$ is query-agnostic (recall from \Cref{sec:model-standard} that we only deal with query-agnostic algorithms in this model), $a_j^{\U}$ and $a_j^{m,\U}$ are conditionally independent given any history and state. In this case, we are interested in independence  conditional on $\F_j^{\U} = \F'$ and $s_j^{\U} = s$. Hence\looseness=-1
\begin{align}
&\ \Pr[\F_{j+1}^{\U} = \F] \nonumber\\
=&\ \Pr[a_j^{\U} =a\mid \F_j^{\U}=\F', s_j^{\U}=s, a_j^{m,\U}=a^m]\ \Pr[s_j^{\U}=s, a_j^{m,\U}=a^m \mid \F_j^{\U} =\F']\ \Pr[\F_j^{\U} =\F']\nonumber\\
=&\ \Pr[a_j^{\U} =a\mid \F_j^{\U}=\F', s_j^{\U}=s]\ \Pr[s_j^{\U}=s, a_j^{m,\U}=a^m \mid \F_j^{\U} =\F']\ \Pr[\F_j^{\U} =\F']\nonumber\\
=&\ \gff^{\A}(\F',s)(a) \cdot  \V^{\U}(\F')(s,a^m)\cdot \Pr[\F_j^{\U} =\F'].\label{eq:part3-full}
\end{align}
By the definition of $\V^{\U}$, we have
\begin{equation}
 \V^{\U}(\F')(s,a^m) =\Pr[s_{t_j}=s,\, a^m_{t_j}=a^m \mid \F_{t_j}\cap\U =\F',\,\q=\U]. \label{eq:part3-adv}
\end{equation}
The definition of $\gk$ implies that $a_{t_j} \sim \gk^{\A}(\F_{t_j}, s_{t_j}) = \gff^{\A}(\F_{t_j}\cap \q, s_{t_j})$. Using the same conditional independence argument as above, we get
\begin{align}
\gff^{\A}(\F',s)(a)=&\ \Pr[a_{t_j} =a\mid \F_{t_j}\cap\U=\F',\, s_{t_j}=s,\, \q=\U]\nonumber\\
=&\ \Pr[a_{t_j} =a\mid \F_{t_j}\cap\U=\F',\, s_{t_j}=s,\, a^m_{t_j}=a^m,\, \q=\U].\label{eq:part3-gamma}
\end{align}
Furthermore, the inductive hypothesis implies that $\Pr[\F_j^{\U} =\F'] = \Pr[\F_{t_j}\cap\U =\F' \mid \q = \U]$. Combining this with Equations~\ref{eq:part3-full}, \ref{eq:part3-adv}, and \ref{eq:part3-gamma} gives us
\begin{align}
&\ \Pr[\F_{j+1}^{\U} = \F] \nonumber\\
=&\ \Pr[a_j^{\U} =a\mid \F_j^{\U}=\F', s_j^{\U}=s, a_j^{m,\U}=a^m]\ \Pr[s_j^{\U}=s, a_j^{m,\U}=a^m \mid \F_j^{\U} =\F']\ \Pr[\F_j^{\U} =\F'] \nonumber\\
=&\  \gff^{\A}(\F',s)(a) \cdot \V^{\U}(\F')(s,a^m) \cdot \Pr[\F_{t_j} = \F' \cap \U \mid \q =\U] \nonumber\\
=&\ \Pr[a_{t_j} =a\mid \F_{t_j} \cap\U=\F',\, s_{t_j}=s,\,a^m_{t_j}=a^m,\, \q=\U] \nonumber \\
& \quad \quad \times \ \Pr[s_{t_j}=s,\, a^m_{t_j}=a^m \mid \F_{t_j}\cap\U =\F',\,\q=\U]\ \Pr[\F_{t_j} \cap \U  =\F'\mid \q=\U] \nonumber\\
=&\ \Pr[\F_{t_j}\cap\U  = \F',\, s_{t_j} = s,\, a_{t_j} =a,\, a^m_{t_j}=a^m \mid \q=\U].\label{eq:part3-main}
\end{align}
By the definition of $t_1,\dots,t_K$, we have $q_{t_j} = 1$. Also, $q_t = 0$ when $t_j < t < t_{j+1}$. Thus
\begin{align}
\F_{t_{j+1}} \cap \q =&\ (s_i, a_i,  a^m_i q_i)_{i \in [t_{j+1}-1]: q_i = 1}\nonumber\\
=&\ (s_i, a_i, a^m_i)_{i \in [t_{j+1}-1]: q_i = 1}\nonumber\\
=&\ \F_{t_j} \cap \q \smallfrown (s_{t_j}, a_{t_j}, a^m_{t_j}). \label{eq:part3-query-res}
\end{align}
Therefore
\begin{align*}
\Pr[\F_{j+1}^{\U} = \F] =&\ \Pr[\F_{t_j}\cap\U  = \F',\, s_{t_j} = s,\, a_{t_j} =a,\, a^m_{t_j}=a^m \mid \q=\U] && (\text{Equation~\ref{eq:part3-main}})\\
=&\ \Pr[\F_{t_j}\cap\q  = \F',\, s_{t_j} = s,\, a_{t_j} =a,\, a^m_{t_j}=a^m \mid \q=\U] && (\text{$\q=\U$})\\\\
=&\ \Pr[(\F_{t_j}\cap \q)  \smallfrown (s_{t_j}, a_{t_j}, a^m_{t_j}) = \F'\smallfrown  (s,a,a^m)\mid \q=\U] && (\text{Rearranging}) \\
=&\ \Pr\big[\F_{t_{j+1}}\cap \q = \F'\smallfrown  (s, a, a^m) \mid \q=\U\big] && (\text{Equation \ref{eq:part3-query-res}})\\
=&\ \Pr[\F_{t_{j+1}}\cap \q = \F \mid \q=\U]. && (\text{Defn of $\F$})
\end{align*}
This completes the induction in Case 2 and thus completes the overall induction, which shows that $\Pr[\F_j^{\U} = \F] =\Pr[\F_{t_j} \cap \U = \F \mid \q = \U]$ for any $\F \in \scrf$ and $j \in [T^{\U}+1]$. Therefore $\D(s_j^{\U}, a_j^{m,\U}, a_j^{\U}) = \D(s_{t_j}, a^m_{t_j}, a_{t_j} \mid \q = \U)$ and thus $\E[g(s_j^{\U}, a_j^{m,\U}, a_j^{\U})] = \E[g(s_{t_j}, a^m_{t_j}, a_{t_j}) \mid \q = \U]$ for any $j \in [T^{\U}]$.
\end{proof}

\thmActive*

\begin{proof}
We have $Q(T,\gk,\V) = \E[\sum_{t=1}^T q_t] = k$ since each $q_t$ is sampled from $\text{Bernoulli}(k/T)$. Thus it remains to bound the regret. For any $\pi \in \Pi$ and $\U \in \{0,1\}^T$, we have
\begin{align*}
\E[\tilD_T(\gk, \pi)\! \mid\! \q\! =\! \U] =&\ \E\left[\sum_{t=1}^T \till_t(a_t, a^m_t) - \sum_{t=1}^T \till_t(\pi(s_t), a^m_t) \mid \q = \U\right] && (\text{Definition of $\tilD_T$})\\
=&\ \frac{T}{k} \E\left[\sum_{t:q_t=1} \ell(a_t, a^m_t) -\sum_{t:q_t=1}\ell(\pi(s_t), a^m_t)  \mid \q = \U\right] && (\text{Definition of $\till_t$})\\
=&\ \frac{T}{k}\E\left[\sum_{j=1}^K \ell(a_{t_j}, a^m_{t_j}) -\sum_{j=1}^K \ell(\pi(s_{t_j}), a^m_{t_j})  \mid \q = \U\right] && (\text{Change of index})\\
=&\ \frac{T}{k}\E\left[\sum_{j=1}^{T^{\U}} \ell(a_{t_j}, a^m_{t_j})-\sum_{j=1}^{T^{\U}} \ell(\pi(s_{t_j}), a^m_{t_j}) \mid \q = \U\right] && (\text{$K = \sum_{t=1}^T q_t = T^{\U}$})\\
=&\ \frac{T}{k}\E\left[\sum_{j=1}^{T^{\U}} \ell(a_j^{\U}, a_j^{m,\U})- \sum_{j=1}^{T^{\U}} \ell(\pi(s_j^{\U}), a_j^{m,\U})\right]. && (\text{Lemma \ref{lem:standard-equiv}})
\end{align*}
Taking a supremum and applying the definition of $\rsa$,
\begin{align*}
\E[\tilD_T(\gk, \pi) \mid \q = \U] \le&\ \sup_{\pi' \in \Pi}\, \E[\tilD_T(\gk, \pi') \mid \q = \U]\\ \le&\ \frac{T}{k} \sup_{\pi' \in \Pi}\, \E\left[\sum_{j=1}^{T^{\U}} \ell(a_j^{\U}, a_j^{m,\U})-\sum_{j=1}^{T^{\U}} \ell(\pi'(s_j^{\U}), a_j^{m,\U}) \right] \\
=&\ \frac{T}{k} \rsa(T^{\U},  \gff,\V^{\U}, \ell). 
\end{align*}
Lemma \ref{lem:smoothness} implies that $\V^{\U}$ is $\sigma$-smooth, so the regret bound of $\gff$ gives us\\$\E[\tilD_T(\gk, \pi) \mid \q = \U] \le \rsa(T^{\U},  \gff,\V^{\U}, \ell) \le R(T^{\U},\sigma)$. By the law of total expectation,
\begin{align*}
\E[\tilD_T(\gk,\pi)] = \E_{\U\sim \D(\q)} \left[\E\big[\tilD_T(\gk,\pi) \mid \q = \U\big]\right] \le \E_{\U\sim \D(\q)}\left[\frac{T}{k} R(T^{\U},\sigma)\right] = \E\left[\frac{T}{k} R(T^{\q},\sigma)\right].
\end{align*}
By definitions, $T^{\q} = \sum_{t=1}^T q_t = K$. Therefore $\E[\tilD_T(\gk,\pi)] \le \frac{T}{k}\E[R(K,\sigma)]$. Now let $R_\sigma(n) = R(n,\sigma)$. Then $R_\sigma$ is concave, so Jensen's inequality implies that
\[
\E[R(K,\sigma)] = \E[R_\sigma(K)] \le R_\sigma(\E[K]) = R(\E[K],\sigma).
\]
Thus $ \E[\tilD_T(\gk,\pi)] \le \frac{T}{k} R(\E[K], \sigma) = \frac{T}{k} R(k,\sigma)$. Since this holds for all $\pi \in \Pi$, we have $\sup_{\pi \in \Pi} \E[\tilD_T(\gk,\pi)] \le \frac{T}{k}R(k,\sigma)$. Combining this with Lemma \ref{lem:unbiased},
\begin{align*}
\rsa(T,\gk,\V,\ell) = \sup_{\pi\in\Pi}\, \E[\Delta_T(\gk,\pi)]
= \sup_{\pi\in\Pi}\, \E[\tilD_T(\gk,\pi)]
\le \frac{T}{k} R(k,\sigma) 
\end{align*}
as required.
\end{proof}

\section{Second Reduction Proofs}\label{sec:ac-to-standard-proof}

This section presents the proofs for our second reduction (\Cref{thm:ac-to-standard}).

\subsection{Proof Notation}

We use the following notation throughout the proof:
\begin{enumerate}[leftmargin=1.5em,
    topsep=0.5ex,
    partopsep=0pt,
    parsep=0pt,
    itemsep=0.5ex]
    \item Fix some $T\in\bbn,\sigma \in [0,1],$ and $\V \in \scrv{\sigma}$.
    \item For each $t \in [T]$, let $X_t$ refer to the value of the variable $X$ in Algorithm~\ref{alg:ac-to-standard} at the start of time step $t$.
    \item We use the variables $\tilde{q}_t$ and $\tilde{a}_t$ as defined in Algorithm~\ref{alg:ac-to-standard}.
    \item Let $M_T = \{t \in [T]: \tilde{a}_t \ne \pi^m(s_t)\}$ be the set of time steps where the action from $\Gamma$ doesn't match the mentor's.
    \item  For each $S \subseteq \s$, let $\vol(S)$ denote the $n$-dimensional Lebesgue measure of $S$.
    \item With slight abuse of notation, we will use inequalities of the form $f(T) \le g(T) + O(h(T))$ to mean that there exists a constant $C$ such that $f(T) \le g(T) + Ch(T)$.
\end{enumerate}

There are two parts of the proof: the regret bounds ($\rmul$ and $\rplus$) and the query bound. To bound the regret, we bound the number of time steps where the agent chooses the wrong action (Lemma \ref{lem:ac-num-errors}) and bound the worst-case reward on the time steps when the agent does choose the wrong action (Lemma \ref{lem:ac-lipschitz-payoff}). Lemma \ref{lem:ac-adv} is an intermediate step which shows the existence of an appropriate $\sigma$-smooth adversary so that we can apply the assumed regret bound on $\Gamma$. Lemma \ref{lem:ac-regret} puts all of the above together to obtain the regret bounds. Lemma \ref{lem:ac-queries} proves the query bound; the main idea is to show that after sufficiently many queries, we will have fully covered the state space.

\subsection{Proof}

\begin{lemma}
\label{lem:ac-adv}
Under the conditions of \Cref{thm:ac-to-standard}, there exists $\V' \in \scrv{\sigma}$ such that\looseness=-1
\begin{align*}
Q(T,\Gamma,\V') =&\ \E\left[\sum_{t=1}^T \tilde{q}_t\right],\\
\rsa(T,\Gamma,\V',\ell_{\bfone}) =&\ \sup_{\pi \in \Pi}\, \E\left[ \sum_{t=1}^T \ell_{\bfone}(\tilde{a}_t,\pi^m(s_t))- \sum_{t=1}^T \ell_{\bfone}(\pi(s_t),\pi^m(s_t))\right].
\end{align*}
\end{lemma}

\begin{proof}
The proof proceeds in three parts. Recall that $\Gamma$ is defined in \Cref{thm:ac-to-standard}.

\textbf{Part 1.} We first claim that for any $t\in[T]$, the variables $(s_i,\tilde{q}_i,\tilde{a}_i)_{i\in[t]}$ and $\pi^m$ uniquely determine the values of $(q_i,a_i,X_i)_{i\in[t]}$. First note that $X_1 = \emptyset$ and for $i \in [t-1]$, $X_{i+1}$ is given by
\[
X_{i+1} = \begin{cases}
X_i \cup \{(s_i,\pi^m(s_i))\} & \text{ if } X_i = \emptyset \text{ or } \min_{(s,a) \in X_i: a = \tilde{a}_i} \norm{s_i - s} > \ep\\
X_i & \text{ otherwise.}
\end{cases}
\]
Thus we can determine the value of $X_{i+1}$ using only $X_i,s_i,\tilde{a}_i,$ and $\pi^m$. Since we know $(s_i,\tilde{q}_i,\tilde{a}_i)_{i\in[t]}, \pi^m,$ and $X_1$, we can determine $X_2$, and use $X_2$ to determine $X_3$, and so on. Once we have determined $X_i$, we can determine $q_i$ and $a_i$ as follows:
\[
(q_i,a_i) = \begin{cases}
(1,\pi^m(s_i)) & \text{ if } X_i = \emptyset \text{ or } \min_{(s,a) \in X_i: a = \tilde{a}_i} \norm{s_i - s} > \ep\\
(\tilde{q}_i, \tilde{a}_i) & \text{ otherwise.}
\end{cases}
\]
Thus $(s_i,\tilde{q}_i,\tilde{a}_i)_{i\in[t]}$ and $\pi^m$ uniquely determine the values of $(q_i,a_i,X_i)_{i\in[t]}$. Then there exists a deterministic function $f: \scrf\to\scrf$ such that $f\big((s_i,\tilde{a}_i, \pi^m(s_i)\tilde{q}_i)_{i\in[t]}\big) = (s_i,a_i, \pi^m(s_i)q_i)_{i\in[t]} = \F_{t+1}$ for any $t \in [T]$. (Recall that $\F_{t+1}$ is the history at the start of time $t+1$ and thus includes $s_t,a_t,q_t$ but not $s_{t+1},a_{t+1},q_{t+1}$.) For a history $\F$, define $\V'(\F) = \V(f(\F))$. Note that the algorithm need not know $f$: since we are defining an adversary, all we need is for this function to exist. Since $\V$ is $\sigma$-smooth by assumption, $\V'$ is also $\sigma$-smooth.

\textbf{Part 2.} We need to show that $\V'$ induces the desired distribution of states and actions. For each $t\in[T]$, let $s_t',q_t',a_t'$ be the random variables corresponding to the state, query decision, and action at time $t$ as induced by $\Gamma$ and $\V'$. Also, for each $t \in [T]$ let $\tilde{\F}_t = \big(s_i,\tilde{a}_i,\pi^m(s_i)\tilde{q}_i\big)_{i\in[t-1]}$ as defined in Algorithm~\ref{alg:ac-to-standard} and let $\F'_t = \big(s_i',a_i',\pi^m(s_i')q_i'\big)_{i\in[t-1]}$. We will show by induction that $\D(\F'_t) = \D(\tilde{\F}_t)$ for each $t \in [T]$. We trivially have $\F'_1 = \emptyset =\tilde{\F}_1$, so assume the claim holds for some $t \in [T]$. Let $\F^+$ be any history of nonzero length and write $\F^+ = \F \smallfrown (s,a,\pi^m(s)q)$ for some $\F \in \scrf,s\in\s,a\in\A,q\in\{0,1\}$. Using definitions and the chain rule of probability,
\begin{align*}
&\ \Pr[\F'_{t+1} = \F^+]\\
=&\ \Pr[\F'_{t+1} = \F \smallfrown (s,a,\pi^m(s)q)] \\
=&\ \Pr[\F'_t = \F, s_t' = s, a_t' = a, q_t' = q] \\
=&\  \Pr[\F'_t = \F]\Pr[s_t' = s \mid \F'_t = \F]\Pr[q_t' = q \mid s_t' = s, \F'_t = \F] \Pr[a_t' = a \mid q_t' = q, s_t' = s, \F'_t = \F].
\end{align*}
By the inductive hypothesis, $\Pr[\F'_t = \F] = \Pr[\tilde{\F}_t = \F]$. Also, by definitions we have $\Pr[s_t' = s \mid \F'_t = \F] = \V'(\F)(s,\pi^m(s))$, $\Pr[q_t' = q \mid s_t' = s, \F'_t = \F]  = \Gamma^Q(\F,s)$, and $\Pr[a_t' = a \mid q_t' = q, s_t' = s, \F'_t = \F] = \Gamma^\A(\F,s,\pi^m(s)q)$. Since $\Gamma$ is query-agnostic by assumption, we have $\Gamma^\A(\F,s,\pi^m(s)q) = \Gamma^\A(\F,s,0)$. Therefore
\begin{align*}
\Pr[\F'_{t+1} = \F^+] =&\ \Pr[\tilde{\F}_t = \F]\boldsymbol{\cdot}\V'(\F)(s,\pi^m(s)) \boldsymbol{\cdot} \Gamma^Q(\F,s)(q)\boldsymbol{\cdot} \Gamma^{\A}(\F,s,\pi^m(s)q)(a) \\
=&\ \Pr[\tilde{\F}_t = \F]\boldsymbol{\cdot}\V(f(\F))(s,\pi^m(s)) \boldsymbol{\cdot}\Gamma^Q(\F,s)(q)\boldsymbol{\cdot} \Gamma^{\A}(\F,s,0)(a).
\end{align*}
Since $s_t \sim \V(\F_t), q_t \sim \Gamma^Q(\F_t,s_t)$, and $\tilde{a}_t \sim \Gamma^{\A}(\F_t,s_t,0)$, we have
\begin{align*}
\Pr[\F'_{t+1} = \F^+] =&\ \Pr[\tilde{\F}_t = \F, s_t = s,\tilde{q}_t = q,\tilde{a}_t = a]\\
=&\ \Pr[\tilde{\F}_t = \F] \boldsymbol{\cdot} \Pr[s_t = s \mid \tilde{\F}_t = \F]  \boldsymbol{\cdot} \Pr[\tilde{q}_t = q \mid s_t = s, \tilde{\F}_t = \F]\\
&\boldsymbol{\cdot} \Pr[\tilde{a}_t = a \mid \tilde{q}_t = q, s_t = s, \tilde{\F}_t = \F]\\
=&\ \Pr[\tilde{\F}_t = \F, s_t = s, \tilde{a}_t = a, \tilde{q}_t = q]\\
=&\ \Pr[\tilde{\F}_{t+1} = \F \smallfrown (s,a,\pi^m(s)q)]\\
=&\ \Pr[\tilde{\F}_{t+1} = \F^+].
\end{align*}
Hence $\D(\F'_{t+1}) = \D(\tilde{\F}_{t+1})$, which completes the induction. In particular, this shows that $\D(q_t') = \D(\tilde{q}_t)$ and $\D(s_t',a_t') = \D(s_t,\tilde{a}_t)$ for all $t \in [T]$.

\textbf{Part 3.} We have
\begin{align*}
Q(T,\Gamma,\V') =&\ \E\left[\sum_{t=1}^T q_t'\right] && (\text{Definition of $Q(T,\Gamma,\V')$ and $q_1',\dots,q_T'$})\\
=&\ \E\left[\sum_{t=1}^T \tilde{q}_t\right]. && (\text{$\D(q_t') = \D(\tilde{q}_t)$ for all $t \in [T]$})
\end{align*}
Similarly, using the definition of $\rsa(T,\Gamma,\V',\ell_{\bfone})$ and $\D(s_t',a_t') = \D(s_t,\tilde{a}_t)$ for all $t \in [T]$,
\begin{align*}
\rsa(T,\Gamma,\V',\ell_{\bfone}) =&\ \sup_{\pi \in \Pi}\, \E\left[\sum_{t=1}^T \ell_{\bfone}(a_t',\pi^m(s_t'))-\sum_{t=1}^T \ell_{\bfone}(\pi(s_t'),\pi^m(s_t'))\right]  \\
=&\ \sup_{\pi \in \Pi}\, \E\left[\sum_{t=1}^T \ell_{\bfone}(\tilde{a}_t,\pi^m(s_t))-\sum_{t=1}^T \ell_{\bfone}(\pi(s_t),\pi^m(s_t)) \right],
\end{align*}
as required.
\end{proof}

\begin{restatable}{lemma}{acNumErrors}
\label{lem:ac-num-errors}
Under the conditions of \Cref{thm:ac-to-standard}, Algorithm~\ref{alg:ac-to-standard} satisfies $\E[|M_T|] \le R(k,T,\sigma)$.
\end{restatable}

\begin{proof}
We have
\begin{align*}
\E[|M_T|] =&\ \E\left[\sum_{t=1}^T \bfone(\tilde{a}_t \ne \pi^m(s_t))\right] && (\text{Definition of $M_T$}) \\
=&\ \E\left[\sum_{t=1}^T \ell_{\bfone}(\tilde{a}_t,\pi^m(s_t))\right] && (\text{Definition of $\ell_{\bfone}$}) \\
=&\ \E\left[ \sum_{t=1}^T \ell_{\bfone}(\tilde{a}_t,\pi^m(s_t))-\sum_{t=1}^T \ell_{\bfone}(\pi^m(s_t),\pi^m(s_t))\right] && (\ell_{\bfone}(\pi^m(s_t),\pi^m(s_t)) = 0)\\
\le&\ \rsa(T,\Gamma, \V',\ell_{\bfone}) && (\text{Lemma~\ref{lem:ac-adv} and $\pi^m \in \Pi$})\\
\le&\ R(k,T,\sigma) && (\text{Assumption of \Cref{thm:ac-to-standard}}),
\end{align*}
as required.
\end{proof}

\begin{lemma}
\label{lem:ac-lipschitz-payoff}
For all $t \in [T]$, $\mu_t(s_t, a_t) \ge \mu_t^m(s_t) - L\ep$.
\end{lemma}
\begin{proof}
Consider an arbitrary $t \in [T]$ and let $\Delta_t = \min_{(s,a) \in X_t: a = \tilde{a}_t} \norm{s_t - s}$. If $\Delta_t > \ep$, then we query the mentor and take action $a_t = \pi^m(s_t)$, so $\mu_t(s_t, a_t) = \mu_t(s_t,\pi^m(s_t)) = \mu_t^m(s_t)$. Now consider $\Delta_t \le \ep$. Let $(s', a') =\argmin_{(s,a) \in X_t: \tilde{a}_t = a} \norm{s_t-s}$; then $\norm{s_t - s'} \le \ep$.

We have $a' = \pi^m(s')$ by construction of $X_t$ and $ a' = \tilde{a}_t$ by construction of $a'$ via the $\argmin$. Combining these with the local generalization assumption, we get
\begin{align*}
\mu_t(s_t, a_t) = \mu_t(s_t, \tilde{a}_t)
= \mu_t(s_t, \pi^m(s'))
\ge \mu_t^m(s_t) - L\norm{s_t - s'}
\ge \mu_t^m(s_t) - L\ep,
\end{align*}
as required.
\end{proof}

\begin{lemma}
    \label{lem:ac-regret}
Under the conditions of \Cref{thm:ac-to-standard}, Algorithm~\ref{alg:ac-to-standard} satisfies
\begin{align*}
\rmul(T,\gac,\V,\bfmu) \le&\ \frac{L\ep R(k,T,\sigma)}{\mum} \quad \text{if } \mum > 0,\\
\rplus(T,\gac,\V,\bfmu) \le&\ L\ep R(k,T,\sigma).
\end{align*}
\end{lemma}

\begin{proof}
First consider any $t\not \in M_T$. Then $\tilde{a}_t = \pi^m(s_t)$, and since $a_t \in \{\tilde{a}_t, \pi^m(s_t)\}$ for all $t$, we have $a_t = \pi^m(s_t)$. Thus $\mu_t(s_t, a_t) = \mu_t^m(s_t)$ for all $t \not\in M_T$. For $t \in M_T$, Lemma~\ref{lem:ac-lipschitz-payoff} implies that $\mu_t^m(s_t) - \mu_t(s_t,a_t) \le L\ep$, so
\begin{align}
\rplus(T,\gac,\V,\bfmu) =&\ \E\left[\sum_{t=1}^T \mu_t(s_t, \pi^m(s_t)) - \sum_{t=1}^T \mu_t(s_t, a_t)\right]\nonumber\\
=&\ \E\left[\sum_{t \in M_T} (\mu_t^m(s_t) - \mu_t(s_t, a_t))\right]\nonumber \\
\le&\ \E\left[\sum_{t \in M_T}L\ep\right]\nonumber \\
=&\ \E[|M_T|] L\ep. \label{eq:rplus-upp}
\end{align}
The analysis for $\rmul$ is similar but requires a bit more work to deal with the product. First, we have
\begin{align*}
\rmul(T, \gac, \V, \bfmu) =&\ \E\left[\log \prod_{t=1}^T \mu_t(s_t, \pi^m(s_t)) - \log \prod_{t=1}^T \mu_t(s_t, a_t)\right]\\
=&\ \E\left[\sum_{t=1}^T \log \frac{\mu_t(s_t, \pi^m(s_t))}{\mu_t(s_t,
a_t)}\right]\\
=&\ \E\left[\sum_{t\in M_T} \log \frac{\mu_t(s_t, \pi^m(s_t))}{\mu_t(s_t,
a_t)}\right].\\
\end{align*}
Note that when $\mum > 0$, $\mu_t(s_t,a_t) \ge \mum > 0$ so the ratio above is well-defined. By Lemma~\ref{lem:ac-lipschitz-payoff},
\begin{align*}
\E\left[\sum_{t\in M_T} \log \frac{\mu_t(s_t, \pi^m(s_t))}{\mu_t(s_t,
a_t)}\right] \le&\ \E\left[\sum_{t\in M_T} \log \frac{\mu_t(s_t,
a_t) + L\ep}{\mu_t(s_t,
a_t)}\right]\\
=&\ \E\left[\sum_{t\in M_T} \log \left(1 + \frac{ L\ep}{\mu_t(s_t,
a_t)}\right)\right]\\
\le&\ \E\left[\sum_{t\in M_T} \log \left(1 + \frac{ L\ep}{\mum}\right)\right].\\
\end{align*}
Since $\log (1+x) \le x$ for all $x \ge 0$, we get
\begin{align}
\rmul(T, \gac, \V, \bfmu) \le&\ \E\left[\sum_{t\in M_T} \log \left(1 + \frac{ L\ep}{\mum}\right)\right]\nonumber\\
\le&\ \E\left[\sum_{t\in M_T} \frac{ L\ep}{\mum}\right]\nonumber\\
=&\ \frac{L\ep\E[|M_T|]}{\mum}. \label{eq:rmul-upp}
\end{align}
Applying Lemma~\ref{lem:ac-num-errors} to Equations~\ref{eq:rplus-upp} and \ref{eq:rmul-upp} completes the proof.
\end{proof}

\begin{definition}[Packing numbers]
\label{def:packing}
Let $(K,\norm{\cdot})$ be a normed vector space and let $\delta > 0$. Then $S\subseteq K$ is a $\delta$-packing of $K$ if for all $a, b \in S$ with $a \ne b$, we have $\norm{a-b} > \delta$. The $\delta$-packing number of $K$, denoted $\M(K,\norm{\cdot}, \delta)$, is the maximum cardinality of any $\delta$-packing of $K$.
\end{definition}

We only consider the Euclidean distance norm, so we just write $M(K, \norm{\cdot}, \delta) = M(K, \delta)$.

\begin{lemma}[Theorem 14.2 in \citealp{wu_lecture_2020}]
    \label{lem:packing}
If $K \subset \bbr^n$ is convex, bounded, and contains a ball with radius $\delta > 0$, then
\[
\M(K,\delta) \le  \frac{3^n \vol(K)}{\delta^n \vol(B)},
\]
where $B$ is a unit ball.
\end{lemma}

\begin{lemma}[Jung's Theorem, \citealp{Jung1901}; Theorem 3.3 in \citealp{gruber2007convex}]
\label{lem:jung}
If $S \subset \bbr^n$ is bounded, then there exists a closed ball with radius at most $\diam(S) \sqrt{\frac{n}{2(n+1)}}$ containing $S$.
\end{lemma}
Jung's Theorem also applies to multisets, since duplicates do not affect the diameter of a set.

\begin{lemma}\label{lem:ac-queries}
Under the conditions of \Cref{thm:ac-to-standard}, Algorithm~\ref{alg:ac-to-standard} satisfies
\[
Q(T,\gac,\V) \in O\left(k + R(k,T,\sigma) + \frac{|\A|\E[\diam(\smols)^n]}{\ep^n}\right).
\]
\end{lemma}

\begin{proof}
Let $\Delta_t = \min_{(s,a) \in X_t: a = \tilde{a}_t} \norm{s_t - s}$. For any $t \in [T]$ with $q_t = 1$, either $\tilde{q}_t = 1$ or $\Delta_t > \ep$ (or both). Thus $
\sum_{t=1}^T q_t \le \sum_{t=1}^T \tilde{q}_t + \sum_{t=1}^T \bfone(\Delta_t > \ep)$. We have $\E[\sum_{t=1}^T \tilde{q}_t] =Q(T,\Gamma,\V')$ by Lemma~\ref{lem:ac-adv} and $Q(T,\Gamma,\V') \le k$ by assumption, so $\E[\sum_{t=1}^T \tilde{q}_t] \le k$. Thus it remains to bound $\E[\sum_{t=1}^T \bfone(\Delta_t > \ep)]$. Let $\hat{Q} = \{t \in [T]: q_t = 1\text{ and  } \Delta_t > \ep\}$. We further subdivide $\hat{Q}$ into $\hat{Q}_1 = \{t\in \hat{Q}: \tilde{a}_t \ne \pi^m(s_t)\}$ and $\hat{Q}_2 = \{t \in \hat{Q}: \tilde{a}_t = \pi^m(s_t)\}$. Since $\hat{Q}_1 \subseteq M_T$, Lemma \ref{lem:ac-num-errors} implies that $ \E[|\hat{Q}_1|] \le  R(k,T,\sigma)$.

Next, fix an $a \in \A$ and let $S_a = \{s \in \smols: \pi^m(s) = a\}$ be the multiset of observed inputs whose mentor action is $a$. Also let $\hats_2 = \{s_t: t \in \hat{Q}_2\}$ be the multiset of inputs associated with time steps in $\hat{Q}_2$. Note that $|\hats_2| = |\hat{Q}_2|$, since $\hats_2$ is a multiset. We claim that $\hats_2 \cap S_a$ is an $\ep$-packing of $S_a$. Suppose instead that there exists $s, s' \in \hats_2 \cap S_a$, with $\norm{s - s'} \le \ep$. WLOG assume $s$ was queried after $s'$ and let $t$ be the time step on which $s$ was queried. Since $s' \in \hats_2$, this implies $(s', \pi^m(s')) \in X_t$. Also, since $s, s' \in \hats_2$ we have $\tilde{a}_t = \pi^m(s_t) = a = \pi^m(s')$. Therefore
\[
\Delta_t = \min_{(x, y) \in X_t: y = \tilde{a}_t} \norm{s_t - x} \le \norm{s_t-s'} \le \ep
\]
which contradicts $t \in \hat{Q}$. Thus $\hats_2 \cap S_a$ is an $\ep$-packing of $S_a$.

By Lemma \ref{lem:jung}, there exists a ball $B_1$ of radius $h:=\diam(\smols)\sqrt{\frac{n}{2(n+1)}}$ which contains $\smols$. Let $B_2$ be the ball with the same center as $B_1$ but with radius $\max(h, \ep)$. Since $S_a \subset \smols \subset B_1 \subset B_2$ and $\hats_2\cap S_a$ is an $\ep$-packing of $S_a$, $\hats_2 \cap S_a$ is also an $\ep$-packing of $B_2$. Also, $B_2$ must contain a ball of radius $\ep$, so Lemma~\ref{lem:packing} implies that 
\begin{align*}
|\hats_2 \cap S_a| \le&\ \M(B_2, \ep)
\le \frac{3^n \vol(B_2)}{\ep^n \vol(B)}
= \big(\max(h, \ep) \big)^n\, \frac{3^n \vol(B)}{\ep^n\vol(B)}\\
=&\ \max\left(\diam(\smols)^n \left(\frac{n}{2(n+1)}\right)^{n/2},\: \ep^n\right) \frac{3^n}{\ep^n}
\ \in\ O\left(\frac{\diam(\smols)^n}{\ep^n} +1\right).
\end{align*}
(The $+1$ is necessary for now since $\diam(\smols)$ could theoretically be zero.) Since $\hats_2 \subseteq \{s_1,\dots,s_T\} \subseteq \cup_{a\in\A}S_a$, we have $|\hats_2| \le \sum_{a \in \A} |\hats_2\cap S_a|$ by the union bound. Therefore 
\begin{align*}
Q(T,\gac,\V) \le&\ \E\left[\sum_{t=1}^T \tilde{q}_t \right] + \E\left[\sum_{t=1}^T \bfone(\Delta_t > \ep)\right]\\
\le&\ k +\E[|\hat{Q}|] \\
=&\ k +\E[|\hat{Q_1}|] + \E[|\hat{Q_2}|] \\
\le&\ k +R(k,T,\sigma) + \E[|\hats_2|] \\
\le&\ k +R(k,T,\sigma)  + \E\left[\sum_{a \in \A} |\hats_2\cap S_a|\right]\\\
\le&\ k +R(k,T,\sigma) + \sum_{a \in \A} O\left(\frac{\E[\diam(\smols)^n]}{\ep^n} +1\right)  \\
\in&\ O\left(k + R(k,T,\sigma) + |\A|\frac{\E[\diam(\smols)^n]}{\ep^n}\right),  \\
\end{align*}
as required.
\end{proof}

\Cref{thm:ac-to-standard} follows from Lemmas~\ref{lem:ac-regret} and \ref{lem:ac-queries}:

\thmACToStandard*

To obtain \Cref{thm:ac-to-standard-final}, we apply \Cref{thm:ac-to-standard} to Corollary \ref{cor:active-to-full}:

\thmACToStandardFinal*

\begin{proof}
By Corollary \ref{cor:active-to-full}, $\gk$ satisfies $\rsa(T,\gk,\V,\ell_{\bfone}) \in O(\frac{T}{k}(d\log k + \sigma^+))$ for any $\sigma \in [0,1]$ and $\V \in \scrv{\sigma}$. Then the conditions of \Cref{thm:ac-to-standard} hold for some $R(k,T,\sigma) \in O(\frac{T}{k}(d\log k + \sigma^+)) = O(T^\frac{1}{2n+2}(d\log T + \sigma^+))$. Thus for any $\V \in \scrv{\sigma}$,
\begin{align*}
\rplus(T, \gac, \V, \bfmu) \in&\ O\left(L \ep T^\frac{1}{2n+2}(d\log T + \sigma^+)\right)\\
=&\ O\left(L T^{\frac{-1}{n+1} + \frac{1}{2n+2}} \left(d\log T + \sigma^+\right)\right)\\
=&\ O\left(L T^\frac{-1}{2n+2} \left(d\log T + \sigma^+\right)\right).
\end{align*}
The bound for $\rmul$ follows from the same arithmetic with an added factor of $1/\mum$. Finally, \Cref{thm:ac-to-standard} also gives us
\begin{align*}
Q(T,\gac,\V) \in&\ O\left(T^{\frac{2n+1}{2n+2}} + T^\frac{1}{2n+2} \left(d\log T + \sigma^+\right) + \frac{\E[\diam(\smols)^n]}{T^\frac{-n}{n+1}}\right)\\
\subseteq&\ O\left(T^\frac{2n+1}{2n+2} (d+\sigma^+ + \E[\diam(\smols)^n])\right),
\end{align*}
as desired.
\end{proof}

We can also apply the reduction to Lemma \ref{lem:ac-num-errors} directly to get the following result, which will be useful later:

\begin{lemma}
    \label{lem:ac-num-errors-final}
Under the conditions of \Cref{thm:ac-to-standard-final}, $\rsa(T,\gac,\V, \ell_{\bfone}) \in O(T^\frac{1}{2n+2}(d\log T + \sigma^+))$ for any $\sigma \in [0,1]$ and $\V \in \scrv{\sigma}$.
\end{lemma}

\begin{proof}
Picking up from the proof of \Cref{thm:ac-to-standard-final}, Lemma \ref{lem:ac-num-errors} gives us $\E[|M_T|] \le R(k,T,\sigma) \in O(T^\frac{1}{2n+2}(d\log T + \sigma^+))$. Then
\begin{align*}
&\ \rsa(T,\gac,\V,\ell_{\bfone})\\
=&\ \sup_{\pi \in \Pi}\, \E\left[\sum_{t=1}^T \ell_{\bfone}(a_t, a_t^m) - \sum_{t=1}^T \ell_{\bfone}(\pi(s_t), a_t^m)\right] && (\text{Definition of $\rsa$})\\
=&\ \sup_{\pi \in \Pi}\, \E\left[\sum_{t=1}^T \ell_{\bfone}(a_t, \pi^m(s_t)) - \sum_{t=1}^T \ell_{\bfone}(\pi(s_t), \pi^m(s_t))\right] && (\text{Assumption of $a_t^m = \pi^m(s_t)$})\\
=&\ \E\left[\sum_{t=1}^T \ell_{\bfone}(a_t, \pi^m(s_t))\right]&& (\text{Assumption of $\pi^m \in \Pi$})\\
\le&\ \E\left[\sum_{t=1}^T \ell_{\bfone}(\tilde{a}_t, \pi^m(s_t))\right]&& (\text{Either $a_t = \tilde{a}_t$ or $a_t = \pi^m(s_t)$})\\
=&\ \E[|M_T|]. && (\text{Definition of $M_T$})\\
\end{align*}
Thus $\rsa(T,\gac,\V,\ell_{\bfone}) \in O(T^\frac{1}{2n+2}(d\log T + \sigma^+))$ as claimed.
\end{proof}

\section{Third Reduction Proofs}\label{sec:mdp-to-ac-proof}

This section presents the proofs for our third reduction (\Cref{thm:mdp-to-ac} and \Cref{thm:mdp-to-ac-no-r}).

\subsection{Proof Notation}

\begin{enumerate}[leftmargin=1.5em,
    topsep=0.5ex,
    partopsep=0pt,
    parsep=0pt,
    itemsep=0.5ex]
    \item Throughout the proof, fix some $ R: \bbn\times[0,1]\to\bbrpos, T \in \bbn, \sigma\in[0,1], (P,\D_1,\pi^m) \in \scrv{\sigma}$ with $(P,\pi^m) \in \p(L)$ and $(r,\pi^m) \in \R(L)$, and some algorithm $\Gamma$.
    \item Let $\V$ denote the adversary defined by $P,\D_1,\pi^m$.
    \item For each $X \subseteq \s$,  let $P(s,a,X)$ denote the probability which $P(s,a)$ assigns to $X$, i.e., $P(s,a,X) =\Pr[s_{t+1} \in X \mid s_t = s, a_t = a]$ for any $t \in [T]$.
    \item Let $p_t$ and $p_t^m$ denote the distributions of $s_t$ and $s_t^m$ respectively. That is, for any $X \subseteq \s$, $p_t(X) = \Pr[s_t\in X]$ and $p_t^m(X)= \Pr[s_t^m \in X]$.
    \item Let $\Delta_t = \sup_{X\subseteq \s}(p_t^m(X) - p_t(X))$.
    \item With ``$N$'' standing for ``next'', for each $X\subseteq \s$ let $N_t(s,X) = \Pr[s_{t+1} \in X \mid s_t = s]$ and $N_t^m(s,X) = \Pr[s_{t+1}^m \in X \mid s_t^m = s]$. 
    \item Let $\alpha_t(X) = \E[N_t^m(s_t, X) - N_t(s_t, X)]$.
\end{enumerate}

\subsection{Proof}

Bounding the action-based regret is a direct application of local generalization of $r$ (Lemma \ref{lem:mdp-action-regret}). It remains to bound the state-based regret. Rather than trying to analyze where the agent's states are better or worse than the mentor's, we simply bound how much $\smols$ and $\sm$ differ at all. The key quantity here is $\Delta_t$. Lemma \ref{lem:mdp-split} shows that a wide range of functions can be bounded using $\Delta_t$. Lemma \ref{lem:mdp-trajectories-induction} provides an initial bound on $\Delta_t$ in terms of a sum of $\alpha_i$'s. Lemma \ref{lem:mdp-trajectories} uses this to show that $\Delta_t \le R(T,\sigma)$. To complete the bound on state-based regret, Lemma \ref{lem:mdp-state-regret} applies Lemma \ref{lem:mdp-split} to the $\Delta_t \le R(T,\sigma)$ bound.

\begin{restatable}[Action-based regret]{lemma}{lemActionRegret}
\label{lem:mdp-action-regret}
Under the conditions of \Cref{thm:mdp-to-ac},\\ $\E [\sum_{t=1}^T r(s_t, \pi^m(s_t)) - \sum_{t=1}^Tr(s_t, a_t)] \le R(T,\sigma)$.
\end{restatable}

\begin{proof}
Define $\mu_t(s,a) = r(s,a)$ for all $t \in [T]$. Since $(r,\pi^m) \in \R(L)$ by assumption, we have $(\bfmu,\pi^m) \in \calU(L)$. Then by assumption of \Cref{thm:mdp-to-ac}, $\E[\sum_{t=1}^T r(s_t, \pi^m(s_t)) - \sum_{t=1}^T r(s_t, a_t)]$$ = \rplus(T,\Gamma,\V, (r,\dots,r))  \le R(T,\sigma)$,
as claimed.
\end{proof}

\begin{restatable}{lemma}{lemSplit}
\label{lem:mdp-split}
For any $t \in [T]$ and any function $f: \s \to [0,1]$,  we have $\E[f(s_t^m) - f(s_t)] \le \Delta_t$.
\end{restatable}

\begin{proof}
By the tail sum formula for expected value (e.g., Lemma 4.4 in \citealp{kallenberg1997foundations}),
\begin{align*}
\E[f(s_t^m)] = \int_{h=0}^\infty \Pr[f(s_t^m) > h] \d h \quad\quad \text{and} \quad\quad \E[f(s_t)] = \int_{h=0}^\infty \Pr[f(s_t) > h] \d h.
\end{align*}
Since $f(s) \in [0,1]$ for all $s \in \s$, we can restrict the integrals to $[0,1]$. Next, for each $h \in \bbrpos$, define $X_h = \{s \in \s: f(s) > h\}$. Then $\Pr[f(s_t^m) > h] = \Pr[s_t^m \in X_h] = p_t^m(X_h)$ and $\Pr[f(s_t) > h] = \Pr[s_t \in X_h] = p_t(X_h)$, and so
\begin{align*}
\E[f(s_t^m) - f(s_t)] =&\ \int_{h=0}^1 \big(\Pr[f(s_t^m) > h] - \Pr[f(s_t) > h]\big) \d h\\
=&\ \int_{h=0}^1 \big(p_t^m(X_h) - p_t(X_h)\big) \d h
\le \int_{h=0}^1 \Delta_t \d h
= \Delta_t.
\end{align*}
\end{proof}

\begin{restatable}{lemma}{lemTrajInd}\label{lem:mdp-trajectories-induction}
For any $t \in [T]$, $\Delta_t \le \sum_{i=1}^{t-1}\sup_{X\subseteq \s}\alpha_i(X)$.
\end{restatable}

\begin{proof}
We proceed by induction. We have $p_1(X) = \D_1(X) = p_1^m(X)$ for all $X \subseteq \s$, so $\Delta_1 = 0 = \sum_{i=1}^0 \sup_{X\subseteq\s} \alpha_i(X)$. Thus the lemma holds for $t=1$.

Now assume the lemma holds for some $t \in [T]$ and fix any $X \subseteq \s$. The next step is to analyze $\E[N_t(s_t, X)]$, which requires a bit of care since $\E[N_t(s_t, X)]$ is an expectation over $s_t$ but $s_t$ also appears in the definition of $N_t(s,X)$. To prevent confusion, we can rewrite $\E[N_t(s_t, X)]$ as $\E_{s\sim \D(s_t)} [N_t(s, X)]$. Then
\begin{align*}
\E[N_t(s_t, X)] =&\ \!\! \E_{s\sim \D(s_t)}[N_t(s, X)]\\
=&\ \!\!\E_{s\sim \D(s_t)}[\Pr[s_{t+1} \in X \mid s_t = s]] \\
=&\ \!\!\E_{s\sim \D(s_t)} \left[\E[\bfone(s_{t+1} \in X) \mid s_t = s]\right].
\end{align*}
By the law of total expectation,
\[
\E_{s\sim \D(s_t)} [\E[\bfone(s_{t+1} \in X) \mid s_t = s]] = \E[\bfone(s_{t+1} \in X)] = \Pr[s_{t+1} \in X].
\]
Therefore $\E[N_t(s_t, X)] = p_{t+1}(X)$. The same argument can be used to show that\\ $\E[N_t^m(s_t^m, X)] = p_{t+1}^m(X)$ by simply replacing $s_t$ and $p_t$ with $s_t^m$ and $p_t^m$ respectively. Thus
\begin{align*}
p_{t+1}^m(X) - p_{t+1}(X) =&\ \E[N_t^m(s_t^m, X)] - \E[N_t(s_t, X)]\\
=&\ \E[N_t^m(s_t^m, X) - N_t^m(s_t, X)] + \E[N_t^m(s_t, X) - N_t(s_t, X)]\\
=&\ \E[N_t^m(s_t^m, X) - N_t^m(s_t, X)] + \alpha_t(X).
\end{align*}
Next, define $f:\s \to[0,1]$ by $f(s) = N_t^m(s,X)$. Then Lemma~\ref{lem:mdp-split} implies that $\E[N_t^m(s_t^m,X) - N_t^m(s_t, X)] \le \Delta_t$, which gives us $p_{t+1}^m(X) - p_{t+1}(X) \le \Delta_t + \alpha_t(X) \le \Delta_t + \sup_{Y\subseteq \s} \alpha_t(Y)$. Since this holds for all $X \subseteq \s$, we have 
\[
\Delta_{t+1} = \sup_{X\subseteq \s} (p_{t+1}^m(X) - p_{t+1}(X)) \le \Delta_t + \sup_{X\subseteq \s} \alpha_t(X).
\]
Combining this with the inductive hypothesis of $\Delta_t \le \sum_{i=1}^{t-1} \sup_{X \subseteq \s} \alpha_i(X)$ gives us
\[
\Delta_{t+1} \le \Delta_t + \sup_{X\subseteq \s} \alpha_t(X) \le \sum_{i=1}^t \sup_{X\subseteq \s} \alpha_i(X)
\]
which completes the induction.
\end{proof}

\begin{restatable}{lemma}{lemTraj}
\label{lem:mdp-trajectories}
If $\Gamma$ satisfies the conditions of \Cref{thm:mdp-to-ac}, then $\Delta_t \le R(T,\sigma)$ for all $t \in [T]$.
\end{restatable}

\begin{proof}
Fix an arbitrary sequence of subsets $X_1,\dots,X_{t-1} \subseteq \s$ and define $\bfmu$ as follows:
\[
\mu_i(s,a) =
\begin{cases}
P(s,a,X_i) & \text{ if } i \in [t-1]\\
1 & \text{ otherwise.}
\end{cases}
\]
For $i > t-1$ we have $|\mu_i(s, \pi^m(s)) - \mu_i(s, \pi^m(s'))| = 0 \le L\norm{s - s'}$. For $i \in [t-1]$, $(P,\pi^m) \in \p(L)$ implies that
\begin{align*}
|\mu_i(s, \pi^m(s)) - \mu_i(s, \pi^m(s'))| =&\ \big|P\big(s,\pi^m(s), X_i)\big) - P(s,\pi^m(s'),X_i)\big)\big|\\
\le&\ \sup_{Y \subseteq \s}\big|P\big(s,\pi^m(s), Y\big) - P(s,\pi^m(s'),Y)\big)\big|\\
=&\ \tv{P(s, \pi^m(s)) - P(s, \pi^m(s'))}\\
\le&\ L \norm{s-s'}.
\end{align*}
Therefore $\bfmu$ satisfies local generalization. Next, we analyze $\mu_i(s_i, a_i)$. For each $i \in [t-1]$,
\begin{align*}
\E[\mu_i(s_i, a_i)] =&\ \E_{s\sim \D(s_i),\, a\sim \D(a_i)}[\mu_i(s, a)]\\
=&\ \E_{s\sim \D(s_i),\, a\sim \D(a_i)}[P(s,a,X_i)] && (\text{Defn of $\mu_i$ for $i \in [t-1]$})\\
=&\ \E_{s\sim \D(s_i),\, a\sim \D(a_i)}[\Pr[s_{i+1} \in X_i \mid s_i = s, a_i = a]] && (\text{Defn of $P(s,a,X_i)$})\\
=&\ \E_{s\sim \D(s_i),\, a\sim \D(a_i)}[\E[\bfone(s_{i+1} \in X_i) \mid s_i = s, a_i = a]] && (\text{Expectation of indicator})\\
=&\ \E[\bfone(s_{i+1} \in X_i)] && (\text{Law of total expectation})\\
=&\ \E[N_i(s_i, X_i)]. && (\text{Proof of Lemma~\ref{lem:mdp-trajectories-induction}})
\end{align*}
Similarly but more easily,
\begin{align*}
N_i^m(s_i,X_i)= \Pr[s_{i+1}^m \in X_i \mid s_i^m = s_i]
= P(s_i, \pi^m(s_i), X_i)
= \mu_i(s_i, \pi^m(s_i)).
\end{align*}
Therefore
\begin{align*}
\sum_{i=1}^{t-1} \alpha_i(X_i) =&\ \sum_{i=1}^{t-1} \E\left[N_i^m(s_i, X_i) - N_i(s_i, X_i)\right]\\
=&\ \sum_{i=1}^{t-1} \E[\mu_i(s_i, \pi^m(s_i)) - \mu_i(s_i, a_i)]\\
=&\ \sum_{i=1}^T \E[\mu_i(s_i, \pi^m(s_i)) - \mu_i(s_i, a_i)]\\
=&\ \rplus(T,\Gamma,\V, \bfmu) \\
\le&\ R(T,\sigma).
\end{align*}
Since this holds for any sequence of subsets $X_1,\dots,X_{t-1} \subseteq \s$, we have
\begin{align*}
\sum_{i=1}^{t-1} \sup_{X\subseteq\s} \alpha_i(X)= \sup_{X_1,\dots,X_{t-1} \subseteq \s} \sum_{i=1}^{t-1} \alpha_i(X_i) \le R(T,\sigma).
\end{align*}
Applying Lemma~\ref{lem:mdp-trajectories-induction} completes the proof.
\end{proof}

\begin{restatable}[State-based regret]{lemma}{lemStateRegret}
\label{lem:mdp-state-regret}
If $\Gamma$ satisfies the conditions of \Cref{thm:mdp-to-ac}, then
\[
\E\left[\sum_{t=1}^T r(s_t^m, \pi^m(s_t^m)) - \sum_{t=1}^T r(s_t, \pi^m(s_t))\right] \le T R(T,\sigma).
\]
\end{restatable}

\begin{proof}
Let $r^m(s) = r(s, \pi^m(s))$ and fix $t \in [T]$. Then Lemma~\ref{lem:mdp-split} implies that $\E[r^m(s_t^m) - r^m(s_t)] \le \Delta_t$. Applying Lemma~\ref{lem:mdp-trajectories} gives us $\E[r^m(s_t^m) - r^m(s_t)] \le R(T,\sigma)$, so
\begin{align*}
\E\left[\sum_{t=1}^T r(s_t^m, \pi^m(s_t^m)) - \sum_{t=1}^T r(s_t, \pi^m(s_t))\right] =&\ \E\left[\sum_{t=1}^T \big(r^m(s_t^m) - r^m(s_t)\big) \right]\\
\le&\ \sum_{t=1}^T R(T,\sigma)\\
=&\ T R(T,\sigma)
\end{align*}
as required.
\end{proof}

\Cref{thm:mdp-to-ac} follows from the bounds on state-based regret and action-based regret (Lemmas~\ref{lem:mdp-action-regret} and \ref{lem:mdp-state-regret}) and the decomposition shown in \Cref{sec:mdp-proof-idea}.

We now move on to \Cref{thm:mdp-to-ac-no-r}, the version of the reduction which does not require local generalization for $r$. As mentioned above, local generalization of $r$ was never used when bounding the state-based regret, so Lemma~\ref{lem:mdp-state-regret} still holds. Thus the only part of the proof that needs to change for \Cref{thm:mdp-to-ac-no-r} is how we bound the action-based regret.

\thmMDPToACNoR*

\begin{proof}
Let $M_T = \{t \in [T]: a_t \ne \pi^m(s_t)\}$. Then
\begin{align*}
\E[|M_T|] =&\ \E\left[\sum_{t=1}^T \bfone(a_t \ne \pi^m(s_t))\right] && (\text{Definition of $M_T$})\\
=&\ \E\left[\sum_{t=1}^T \ell_{\bfone}(a_t ,\pi^m(s_t))- \sum_{t=1}^T \ell_{\bfone}(\pi^m(s_t),\pi^m(s_t))\right] && (\text{Definition of $\ell_{\bfone}$})\\
\le&\ \sup_{\pi \in \Pi}\, \E\left[\sum_{t=1}^T \ell_{\bfone}(a_t ,\pi^m(s_t))- \sum_{t=1}^T \ell_{\bfone}(\pi(s_t),\pi^m(s_t))\right] && (\text{$\pi^m \in \Pi$})\\
=&\ \rsa(T,\Gamma,\V,\ell_{\bfone}) && (\text{Definition of $\rsa$})\\
\le&\ R_1(T,\sigma). && (\text{Theorem assumption})
\end{align*}
Since $r(s_t,\pi^m(s_t)) = r(s_t,a_t)$ for all $t \not\in M_T$ and $r(s,a) \in [0,1]$ for all $s\in \s,a\in \A$, we have
\begin{align*}
\E \left[\sum_{t=1}^T r(s_t, \pi^m(s_t)) - \sum_{t=1}^T r(s_t, a_t)\right]  =&\ \E \left[\sum_{t\in M_T}\big( r(s_t, \pi^m(s_t)) - r(s_t, a_t)\big)\right]\\
\le&\ \E \left[\sum_{t\in M_T}1 \right]
= \E[|M_T|]
\le R_1(T,\sigma).
\end{align*}
Thus the action-based regret is at most $R_1(T,\sigma)$. Lemma~\ref{lem:mdp-state-regret} tells us that the state-based regret is at most $T R_2(T,\sigma)$. Combining these proves the theorem.
\end{proof}

Lastly, we apply \Cref{thm:mdp-to-ac-no-r} to Corollary~\ref{cor:active-to-full} to obtain our final result.

\thmMDPToACFinal*

\begin{proof}
\Cref{thm:ac-to-standard-final} gives us the required bounds on $\rplus(T,\gac,\V,\bfmu)$ and $Q(T,\gac,\V) = Q(T,\gac,(P,\D_1,\pi^m))$. Lemma \ref{lem:ac-num-errors-final} gives us the required bound on $\rsa(T,\gac,\V,\ell_{\bfone})$. Thus we can apply \Cref{thm:mdp-to-ac-no-r} with some $R_1(T,\sigma) \in  O(T^\frac{1}{2n+2} (d\log T + \sigma^+))$ and $R_2(T,\sigma) \in O(L T^\frac{-1}{2n+2} (d\log T + \sigma^+))$ to obtain
\begin{align*}
\rmdp(T,\Gamma,(P,\D_1,\pi^m), r) \in&\ O\left(T^\frac{1}{2n+2} (d\log T + \sigma^+) + T \cdot L T^\frac{-1}{2n+2} (d\log T + \sigma^+)\right)\\
=&\ O\left(LT^\frac{2n+1}{2n+2}(d\log T + \sigma^+)\right)
\end{align*}
as desired.
\end{proof}

\section{Logarithmic Regret in Full-Feedback with Realizability}\label{sec:realizable-alg}

\Cref{thm:active-to-full} reduces active online learning to full-feedback online learning. However, to obtain our desired subconstant regret, we will need a full-feedback algorithm with a better regret bound than the typical $O(\sqrt{T})$ bound. Lemma~\ref{lem:littlestone} provides such a bound under the assumptions of $\pi^m \in \Pi$ (also known as \emph{realizability}) and finite Littlestone dimension. However, finite Littlestone dimension is quite a strong assumption: even some simple classes have infinite Littlestone dimension (for example, the class of thresholds on $[0,1]$: see Example 21.4 in \citealp{shalev-shwartz_understanding_2014}). In this section, we provide an algorithm with $O(\log T)$ regret for $\sigma$-smooth adversaries under the weaker assumption of finite VC dimension (also assuming realizability).

Our algorithm involves approximating $\Pi$ by a finite class $\tilpi$. Formally: for $\ep > 0$, a policy class $\tilpi$ is an \emph{$\ep$-cover} of a policy class $\Pi$ if for every $\pi \in \Pi$, there exists $\stilpi \in \tilpi$ such that $\Pr_{s\sim \beta} [\pi(s) \ne \stilpi(s)] \le \ep$. 
The existence of small $\ep$-covers is crucial:

\begin{lemma}
\label{lem:smooth-cover}
For all $\ep > 0$, any policy class of VC dimension $d$ admits an $\ep$-cover of size at most $(41 /\ep)^d$.
\end{lemma}

There are many references for bounds of the form in Lemma~\ref{lem:smooth-cover}. Lemma 7.3.2 in \cite{haghtalab2018foundation} is one (although it nominally assumes that $\beta$ is the uniform distribution, the exact same proof holds for an arbitrary $\beta$). See also \citet{haussler_generalization_1995} or Lemma 13.6 in \citet{boucheron2013concentration} for similar bounds.

We next show that an $\ep$-cover is a good approximation for any $\sigma$-smooth distribution:


\begin{lemma}
\label{lem:smooth-concentrate}
Let $\tilpi$ be an $\ep$-cover of $\Pi$ and let $\sigma \in (0,1]$. Then for any $\pi \in \Pi$, there exists $\stilpi \in \tilpi$ such that for any $\sigma$-smooth distribution $\D$,  $\Pr_{s\sim \D}[\pi(s) \ne \stilpi(s)] \le \ep/\sigma$.
\end{lemma}

\begin{proof}
Define $S(\stilpi) = \{s \in \s: \pi(s) \ne \stilpi(s)\}$. By the definition of an $\ep$-cover, there exists $\stilpi \in \tilpi$ such that $\Pr_{s\sim \beta}[s \in S(\stilpi)] \le \ep$. Since $\D$ is $\sigma$-smooth, $\Pr_{s \sim \D}[\pi(s) \ne \stilpi(s)] =\Pr_{s\sim \D}[s \in S(\stilpi)] \le \Pr_{s\sim \beta}[s \in S(\stilpi)]/\sigma \le \ep/\sigma$, as claimed.
\end{proof}

Our last ingredient is Lemma~\ref{lem:small-loss}, which provides a particular type of loss bound:

\begin{lemma}[Corollary 2.3 in \citealp{cesa2006prediction}]
\label{lem:small-loss}
There exists an algorithm $\Gamma$ such that for any adversary, when $\Gamma$ is run on a finite policy class $\Pi$, we have
\[
\E\left[\sum_{t=1}^T \ell_{\bfone}(a_t,a_t^m)\right] \le \frac{e}{e-1} \left( \min_{\pi \in \Pi} \E\left[\sum_{t=1}^T \ell_{\bfone}(\pi(s_t), a_t^m)\right] + \ln |\Pi|\right).
\]
\end{lemma}

We can now prove the main result of this section. The algorithm needs to know $T$ (in order to construct an appropriate $\ep$-cover) but not $\sigma$.

\lemRealizableAlg*

\begin{proof}
By Lemma~\ref{lem:smooth-cover}, $\Pi$ admits a $\frac{1}{T}$-cover of size at most $(41 T)^d$; let $\tilpi$ be any such $\frac{1}{T}$-cover. Let $\Gamma$ be the algorithm from Lemma~\ref{lem:small-loss} and define $\gff$ by running $\Gamma$ on $\tilpi$. Letting $c = \frac{e}{e-1}$,
\begin{align*}
\rsa(T,\gff,\V,\ell_{\bfone}) =&\ \sup_{\pi \in \Pi}\, \E\left[ \sum_{t=1}^T \ell_{\bfone}(a_t, a_t^m) - \sum_{t=1}^T \ell_{\bfone}(\pi(s_t), a_t^m)\right]\\
=&\ \E\left[ \sum_{t=1}^T \ell_{\bfone}(a_t, \pi^m(s_t)) \right]\\
\le&\  c \left( \min_{\stilpi \in \tilpi} \E\left[\sum_{t=1}^T \ell_{\bfone}(\stilpi(s_t), \pi^m(s_t))\right] + \ln |\tilpi|\right)\\
\le&\  c \left( \min_{\stilpi \in \tilpi} \sum_{t=1}^T \Pr[\stilpi(s_t) \ne \pi^m(s_t)] + \ln |\tilpi|\right).
\end{align*}
Since $\V$ is $\sigma$-smooth, we have $\Pr_{(s,a)\sim \V(\F)} [s \in X] \le \beta(X) /\sigma$ for any $X\subseteq \s$ and $\F \in \scrf$. Then by the law of total expectation, for any $X \subseteq \s$ we have
\begin{align*}
\Pr[s_t \in X] = \E_{\F_t} [\Pr[s_t \in X  \mid \F_t]]
= \E_{\F_t} \left[\Pr_{(s,a)\sim \V(\F_t)}[s \in X]\right]
\le \E_{\F_t} \left[\frac{\beta(X)}{\sigma}\right]
\le \frac{\beta(X)}{\sigma}.
\end{align*}
Thus the distribution of $s_t$ is also $\sigma$-smooth, so by Lemma~\ref{lem:smooth-concentrate}, 
\begin{align*}
\rsa(T,\gff,\V,\ell_{\bfone}) \le&\ c\left( \min_{\stilpi \in \tilpi} \sum_{t=1}^T \Pr[\stilpi(s_t) \ne \pi^m(s_t)] + \ln |\tilpi|\right)\\
\le c \left(  \sum_{t=1}^T \frac{1}{T\sigma} + \ln |\tilpi|\right)&
= c \left( \sigma^{-1} + \ln |\tilpi|\right)
= c \left(\sigma^{-1} + \ln (41 T)^d \right)
\in O(d \log T + \sigma^+)
\end{align*}
as required.
\end{proof}

\bibliography{refs}

@incollection{Abaimov2020,
  title = {Artificial Intelligence in Autonomous Weapon Systems},
  booktitle = {21st Century Prometheus: {{Managing CBRN}} Safety and Security Affected by Cutting-Edge Technologies},
  author = {Abaimov, Stanislav and Martellini, Maurizio},
  editor = {Martellini, Maurizio and Trapp, Ralf},
  year = 2020,
  pages = {141--177},
  publisher = {Springer International Publishing},
  address = {Cham}
}

@book{altman1999constrained,
  title = {Constrained Markov Decision Processes},
  author = {Altman, Eitan},
  year = 1999,
  volume = {7},
  publisher = {CRC Press}
}

@inproceedings{azar_minimax_2017,
  title = {Minimax Regret Bounds for Reinforcement Learning},
  booktitle = {International Conference on Machine Learning},
  author = {Azar, Mohammad Gheshlaghi and Osband, Ian and Munos, R{\'e}mi},
  year = 2017,
  month_not_used = jul,
  pages = {263--272},
  publisher_not_used = {PMLR},
  urldate = {2024-01-13},
  langid = {english}
}

@article{bengio2024managing,
  title = {Managing Extreme {{AI}} Risks amid Rapid Progress},
  author = {Bengio, Yoshua and Hinton, Geoffrey and Yao, Andrew and Song, Dawn and Abbeel, Pieter and Darrell, Trevor and Harari, Yuval Noah and Zhang, Ya-Qin and Xue, Lan and {Shalev-Shwartz}, Shai and others},
  year = 2024,
  journal = {Science},
  volume = {384},
  number = {6698},
  pages = {842--845},
  publisher = {American Association for the Advancement of Science}
}

@article{besson2018doubling,
  title = {What Doubling Tricks Can and Can't Do for Multi-Armed Bandits},
  author = {Besson, Lilian and Kaufmann, Emilie},
  year = 2018,
  journal = {arXiv preprint 1803.06971},
  eprint = {1803.06971},
  archiveprefix = {arXiv}
}

@inproceedings{block2022smoothed,
  title = {Smoothed Online Learning Is as Easy as Statistical Learning},
  booktitle = {Conference on Learning Theory},
  author = {Block, Adam and Dagan, Yuval and Golowich, Noah and Rakhlin, Alexander},
  year = 2022,
  pages = {1716--1786},
  publisher_not_used = {PMLR}
}

@book{boucheron2013concentration,
  title = {Concentration Inequalities: A Nonasymptotic Theory of Independence},
  author = {Boucheron, St{\'e}phane and Lugosi, G{\'a}bor and Massart, Pascal},
  year = 2013,
  month_not_used = feb,
  publisher = {Oxford University Press}
}

@article{cesa2005minimizing,
  title = {Minimizing Regret with Label Efficient Prediction},
  author = {{Cesa-Bianchi}, Nicolo and Lugosi, G{\'a}bor and Stoltz, Gilles},
  year = 2005,
  journal = {IEEE Transactions on Information Theory},
  volume = {51},
  number = {6},
  pages = {2152--2162},
  publisher = {IEEE}
}

@book{cesa2006prediction,
  title = {Prediction, Learning, and Games},
  author = {{Cesa-Bianchi}, Nicolo and Lugosi, G{\'a}bor},
  year = 2006,
  publisher = {Cambridge University Press}
}

@article{cohen_curiosity_2021,
  title = {Curiosity Killed or Incapacitated the Cat and the Asymptotically Optimal Agent},
  author = {Cohen, Michael K and Catt, Elliot and Hutter, Marcus},
  year = 2021,
  month_not_used = jun,
  journal = {IEEE Journal on Selected Areas in Information Theory},
  volume = {2},
  number = {2},
  pages = {665--677},
  urldate = {2024-01-13}
}

@inproceedings{cohen_pessimism_2020,
  title = {Pessimism about unknown unknowns inspires conservatism},
  booktitle = {{{Conference}} on {{Learning Theory}}},
  author = {Cohen, Michael K and Hutter, Marcus},
  year = 2020,
  month_not_used = jul,
  pages = {1344--1373},
  publisher_not_used = {PMLR},
  urldate = {2024-01-13},
  langid = {english}
}

@inproceedings{cortes2018online,
  title = {Online Learning with Abstention},
  booktitle = {International Conference on Machine Learning},
  author = {Cortes, Corinna and DeSalvo, Giulia and Gentile, Claudio and Mohri, Mehryar and Yang, Scott},
  year = 2018,
  pages = {1059--1067},
  publisher_not_used = {PMLR}
}

@article{critch2023tasra,
  title = {{{TASRA}}: A Taxonomy and Analysis of Societal-Scale Risks from {{AI}}},
  author = {Critch, Andrew and Russell, Stuart},
  year = 2023,
  journal = {arXiv preprint 2306.06924},
  eprint = {2306.06924},
  archiveprefix = {arXiv}
}

@article{daniely2015multiclass,
  title = {Multiclass Learnability and the {{ERM}} Principle},
  author = {Daniely, Amit and Sabato, Sivan and {Ben-David}, Shai and {Shalev-Shwartz}, Shai},
  year = 2015,
  journal = {Journal of Machine Learning Research},
  volume = {16},
  number = {1},
  pages = {2377--2404},
  publisher = {JMLR}
}

@article{esser_actioneffect_2023,
  title = {Action--Effect Knowledge Transfers to Similar Effect Stimuli},
  author = {Esser, Sarah and Haider, Hilde and Lustig, Clarissa and Tanaka, Takumi and Tanaka, Kanji},
  year = 2023,
  month_not_used = oct,
  journal = {Psychological Research},
  volume = {87},
  number = {7},
  pages = {2249--2258},
  urldate = {2024-11-17},
  langid = {english}
}

@article{garcia_comprehensive_2015,
  title = {A Comprehensive Survey on Safe Reinforcement Learning},
  author = {Garc{\'i}a, Javier and Fern{\'a}ndez, Fernando},
  year = 2015,
  journal = {Journal of Machine Learning Research},
  volume = {16},
  number = {42},
  pages = {1437--1480},
  urldate = {2024-01-13}
}

@book{gruber2007convex,
  title = {Convex and Discrete Geometry},
  author = {Gruber, Peter M},
  year = 2007,
  publisher = {Springer}
}

@article{gu_review_2024,
  title = {A Review of Safe Reinforcement Learning: {{Methods}}, Theories, and Applications},
  author = {Gu, Shangding and Yang, Long and Du, Yali and Chen, Guang and Walter, Florian and Wang, Jun and Knoll, Alois},
  year = 2024,
  journal = {IEEE Transactions on Pattern Analysis and Machine Intelligence},
  volume = {46},
  number = {12},
  pages = {11216--11235},
  publisher = {IEEE}
}

@article{guembe_emerging_2022,
  title = {The Emerging Threat of {AI}-driven Cyber Attacks: A Review},
  shorttitle = {The {{Emerging Threat}} of {{AI-driven Cyber Attacks}}},
  author = {Guembe, Blessing and Azeta, Ambrose and Misra, Sanjay and Osamor, Victor Chukwudi and {Fernandez-Sanz}, Luis and Pospelova, Vera},
  year = 2022,
  month_not_used = dec,
  journal = {Applied Artificial Intelligence},
  volume = {36},
  number = {1},
  urldate = {2024-09-04},
  langid = {english}
}

@inproceedings{hadfield-menell_inverse_2017,
  title = {Inverse Reward Design},
  booktitle = {Conference on Neural Information Processing Systems},
  author = {{Hadfield-Menell}, Dylan and Milli, Smitha and Abbeel, Pieter and Russell, Stuart and Dragan, Anca D.},
  year = 2017,
  month_not_used = dec,
  pages = {6768--6777},
}

@phdthesis{haghtalab2018foundation,
  title = {Foundation of Machine Learning, by the People, for the People},
  author = {Haghtalab, Nika},
  year = 2018,
  school = {Carnegie Mellon University}
}

@inproceedings{haghtalab2022oracle,
  title = {Oracle-Efficient Online Learning for Smoothed Adversaries},
  booktitle = {Conference on Neural Information Processing Systems},
  author = {Haghtalab, Nika and Han, Yanjun and Shetty, Abhishek and Yang, Kunhe},
  year = 2022,
  volume = {35},
  pages = {4072--4084},
  publisher = {Curran Associates, Inc.}
}

@article{haghtalab2024smoothed,
  title = {Smoothed Analysis with Adaptive Adversaries},
  author = {Haghtalab, Nika and Roughgarden, Tim and Shetty, Abhishek},
  year = 2024,
  journal = {Journal of the ACM},
  volume = {71},
  number = {3},
  pages = {1--34},
  publisher = {ACM New York, NY}
}

@article{hajian_transfer_2019,
  title = {Transfer of Learning and Teaching: A Review of Transfer Theories and Effective Instructional Practices},
  shorttitle = {Transfer of {{Learning}} and {{Teaching}}},
  author = {Hajian, Shiva},
  year = 2019,
  journal = {IAFOR Journal of Education},
  volume = {7},
  number = {1},
  pages = {93--111},
  urldate = {2024-11-17},
  langid = {english}
}

@article{hanneke2014active,
  title = {Theory of Disagreement-Based Active Learning},
  author = {Hanneke, Steve},
  year = 2014,
  journal = {Foundations and Trends\textregistered{} in Machine Learning},
  volume = {7},
  number = {2-3},
  pages = {131--309},
  publisher = {Now Publishers, Inc.}
}

@article{haussler_generalization_1995,
  title = {A Generalization of {{Sauer}}'s Lemma},
  author = {Haussler, David and Long, Philip M},
  year = 1995,
  month_not_used = aug,
  journal = {Journal of Combinatorial Theory, Series A},
  volume = {71},
  number = {2},
  pages = {219--240},
  urldate = {2024-05-06}
}

@inproceedings{he2021nearly,
  title = {Nearly Minimax Optimal Reinforcement Learning for Discounted {{MDPs}}},
  author = {He, Jiafan and Zhou, Dongruo and Gu, Quanquan},
  year = 2021,
  booktitle = {Conference on Neural Information Processing Systems},
  volume = {34},
  pages = {22288--22300}
}

@article{hendrycks2023overviewcatastrophicairisks,
  title = {An Overview of Catastrophic {{AI}} Risks},
  author = {Hendrycks, Dan and Mazeika, Mantas and Woodside, Thomas},
  year = 2023,
  journal = {arXiv preprint 2306.12001},
  eprint = {2306.12001},
  archiveprefix = {arXiv}
}

@article{jaksch_near-optimal_2010,
  title = {Near-Optimal Regret Bounds for Reinforcement Learning},
  author = {Jaksch, Thomas and Ortner, Ronald and Auer, Peter},
  year = 2010,
  journal = {Journal of Machine Learning Research},
  volume = {11},
  number = {51},
  pages = {1563--1600},
  urldate = {2024-01-13}
}

@article{Jung1901,
  title = {Ueber Die Kleinste {{Kugel}}, Die Eine R\"aumliche {{Figur}} Einschliesst.},
  author = {Jung, Heinrich},
  year = 1901,
  journal = {Journal f\"ur die reine und angewandte Mathematik},
  volume = {123},
  pages = {241--257}
}

@book{kallenberg1997foundations,
  title = {Foundations of Modern Probability},
  author = {Kallenberg, Olav},
  year = 1997,
  publisher = {Springer}
}

@inproceedings{kohli2020enabling,
  title = {Enabling Pedestrian Safety Using Computer Vision Techniques: A Case Study of the 2018 {{Uber Inc}}. Self-Driving Car Crash},
  booktitle = {Future of Information and Communication Conference},
  author = {Kohli, Puneet and Chadha, Anjali},
  year = 2019,
  pages = {261--279},
  publisher = {Springer}
}

@article{kosoy_delegative_2019,
  title = {Delegative Reinforcement Learning: Learning to Avoid Traps with a Little Help},
  shorttitle = {Delegative {{Reinforcement Learning}}},
  author = {Kosoy, Vanessa},
  year = 2019,
  month_not_used = jul,
  journal = {arXiv preprint 1907.08461},
  urldate = {2024-01-13}
}

@article{krasowski23,
  title = {Provably Safe Reinforcement Learning: Conceptual Analysis, Survey, and Benchmarking},
  author = {Krasowski, Hanna and Thumm, Jakob and M{\"u}ller, Marlon and Sch{\"a}fer, Lukas and Wang, Xiao and Althoff, Matthias},
  year = 2023,
  journal = {Transactions on Machine Learning Research},
  langid = {english}
}

@inproceedings{lattimore_asymptotically_2011,
  title = {Asymptotically Optimal Agents},
  booktitle = {International Conference on Algorithmic Learning Theory},
  author = {Lattimore, Tor and Hutter, Marcus},
  year = 2011,
  month_not_used = oct,
  pages = {368--382},
  publisher = {Springer-Verlag},
  address = {Berlin, Heidelberg},
  urldate = {2024-11-19}
}

@inproceedings{li2008knows,
  title = {Knows What It Knows: a Framework for Self-Aware Learning},
  booktitle = {International Conference on Machine Learning},
  author = {Li, Lihong and Littman, Michael L and Walsh, Thomas J},
  year = 2008,
  pages = {568--575}
}

@article{littlestone1988learning,
  title = {Learning Quickly When Irrelevant Attributes Abound: a New Linear-Threshold Algorithm},
  author = {Littlestone, Nick},
  year = 1988,
  journal = {Machine Learning},
  volume = {2},
  pages = {285--318},
  publisher = {Springer}
}

@article{liu_regret_2021,
  title = {Regret bounds for discounted {MDPs}},
  author = {Liu, Shuang and Su, Hao},
  year = 2021,
  month_not_used = may,
  journal = {arXiv preprint 2002.05138},
  urldate = {2024-11-19}
}

@inproceedings{liu2021learning,
  title = {Learning Policies with Zero or Bounded Constraint Violation for Constrained {{MDPs}}},
  author = {Liu, Tao and Zhou, Ruida and Kalathil, Dileep and Kumar, Panganamala and Tian, Chao},
  year = 2021,
  booktitle = {Conference on Neural Information Processing Systems},
  volume = {34},
  pages = {17183--17193}
}

@inproceedings{maillard_active_2019,
  title = {Active roll-outs in {MDPs} with irreversible dynamics},
  author = {Maillard, Odalric-Ambrym and Mann, Timothy and Ortner, Ronald and Mannor, Shie},
  publisher = {HAL Open Science},
  year = 2019,
  month_not_used = jul,
  urldate = {2024-01-15}
}

@inproceedings{model_zhao_22a,
  title = {Model-Free Safe Control for Zero-Violation Reinforcement Learning},
  booktitle = {Conference on Robot Learning},
  author = {Zhao, Weiye and He, Tairan and Liu, Changliu},
  editor = {Faust, Aleksandra and Hsu, David and Neumann, Gerhard},
  year = 2022,
  month_not_used = nov,
  volume = {164},
  pages = {784--793},
  publisher_not_used = {PMLR}
}

@techreport{mouton2024operational,
  title = {The Operational Risks of {{AI}} in Large-Scale Biological Attacks},
  author = {Mouton, Christopher A and Lucas, Caleb and Guest, Ella},
  year = 2024,
  institution = {RAND Corporation, Santa Monica}
}

@article{natarajan_learning_1989,
  title = {On Learning Sets and Functions},
  author = {Natarajan, Balas Kausik},
  year = 1989,
  month_not_used = oct,
  journal = {Machine Learning},
  volume = {4},
  number = {1},
  pages = {67--97},
  urldate = {2024-05-06},
  langid = {english}
}

@techreport{national_institute_of_standards_and_technology_us_artificial_2024,
  title = {Artificial Intelligence Risk Management Framework: Generative Artificial Intelligence Profile},
  shorttitle = {Artificial Intelligence Risk Management Framework},
  author = {{National Institute of Standards and Technology}},
  year = 2024,
  month_not_used = jul,
  address = {Gaithersburg, MD},
  institution = {{National Institute of Standards and Technology}},
  urldate = {2024-12-09},
  langid = {english}
}

@inproceedings{ortner2012online,
  title = {Online Regret Bounds for Undiscounted Continuous Reinforcement Learning},
  author = {Ortner, Ronald and Ryabko, Daniil},
  year = 2012,
  booktitle = {Conference on Neural Information Processing Systems},
  volume = {25}
}

@article{osa_algorithmic_2018,
  title = {An Algorithmic Perspective on Imitation Learning},
  author = {Osa, Takayuki and Pajarinen, Joni and Neumann, Gerhard and Bagnell, J Andrew and Abbeel, Pieter and Peters, Jan},
  year = 2018,
  journal = {Foundations and Trends\textregistered{} in Robotics},
  volume = {7},
  number = {1-2},
  pages = {1--179},
  publisher = {Emerald Publishing Limited}
}

@inproceedings{plaut2025avoiding,
  title = {Avoiding Catastrophe in Online Learning by Asking for Help},
  booktitle = {International Conference on Machine Learning},
  author = {Plaut, Benjamin and Zhu, Hanlin and Russell, Stuart},
  year = 2025,
  month_not_used = jul,
  volume = {267},
  pages = {49537--49568},
  publisher_not_used = {PMLR}
}

@book{puterman2014markov,
  title = {Markov Decision Processes: Discrete Stochastic Dynamic Programming},
  author = {Puterman, Martin L},
  year = 2014,
  publisher = {John Wiley \& Sons}
}

@book{quinonero2022dataset,
  title = {Dataset Shift in Machine Learning},
  author = {{Qui{\~n}onero-Candela}, Joaquin and Sugiyama, Masashi and Schwaighofer, Anton and Lawrence, Neil D},
  year = 2022,
  publisher = {MIT Press}
}

@article{rajpurkar2022ai,
  title = {{AI} in Health and Medicine},
  author = {Rajpurkar, Pranav and Chen, Emma and Banerjee, Oishi and Topol, Eric J},
  year = 2022,
  journal = {Nature Medicine},
  volume = {28},
  number = {1},
  pages = {31--38},
  publisher = {Nature Publishing Group US New York}
}

@book{shalev-shwartz_understanding_2014,
  title = {Understanding {{Machine Learning}}: {{From Theory}} to {{Algorithms}}},
  shorttitle = {Understanding {{Machine Learning}}},
  author = {{Shalev-Shwartz}, Shai and {Ben-David}, Shai},
  year = 2014,
  month_not_used = may,
  edition = {1},
  publisher = {Cambridge University Press},
  urldate = {2024-05-06},
  langid = {english}
}

@article{slattery2024ai,
  title = {The {{AI}} Risk Repository: A Comprehensive Meta-Review, Database, and Taxonomy of Risks from Artificial Intelligence},
  author = {Slattery, Peter and Saeri, Alexander K and Grundy, Emily AC and Graham, Jess and Noetel, Michael and Uuk, Risto and Dao, James and Pour, Soroush and Casper, Stephen and Thompson, Neil},
  year = 2024,
  journal = {arXiv preprint 2408.12622},
  eprint = {2408.12622},
  archiveprefix = {arXiv}
}

@article{stradi2024learning,
  title = {Learning Adversarial {{MDPs}} with Stochastic Hard Constraints},
  author = {Stradi, Francesco Emanuele and Castiglioni, Matteo and Marchesi, Alberto and Gatti, Nicola},
  year = 2024,
  journal = {arXiv preprint 2403.03672},
  eprint = {2403.03672},
  archiveprefix = {arXiv}
}

@article{villasenor2020artificial,
  title = {Artificial Intelligence, Due Process and Criminal Sentencing},
  author = {Villasenor, John and Foggo, Virginia},
  year = 2020,
  journal = {Michigan State Law Review},
  pages = {295--354},
  publisher = {HeinOnline}
}

@techreport{wu_lecture_2020,
  title = {Lecture Notes on: {{Information-theoretic}} Methods for High-Dimensional Statistics},
  author = {Wu, Yihong},
  year = 2020,
  institution = {Yale University}
}

@article{yang2024generalized,
  title = {Generalized Out-of-Distribution Detection: a Survey},
  author = {Yang, Jingkang and Zhou, Kaiyang and Li, Yixuan and Liu, Ziwei},
  year = 2024,
  journal = {International Journal of Computer Vision},
  volume = {132},
  number = {12},
  pages = {5635--5662},
  publisher = {Springer}
}

@inproceedings{zhang2020almost,
  title = {Almost Optimal Model-Free Reinforcement Learning via Reference-Advantage Decomposition},
  author = {Zhang, Zihan and Zhou, Yuan and Ji, Xiangyang},
  year = 2020,
  booktitle = {Conference on Neural Information Processing Systems},
  volume = {33},
  pages = {15198--15207}
}

@inproceedings{zhao_state-wise_2023,
  title = {State-Wise Safe Reinforcement Learning: a Survey},
  shorttitle = {State-Wise {{Safe Reinforcement Learning}}},
  booktitle = {International Joint Conference on Artificial Intelligence},
  author = {Zhao, Weiye and He, Tairan and Chen, Rui and Wei, Tianhao and Liu, Changliu},
  year = 2023,
  month_not_used = aug,
  volume = {6},
  pages = {6814--6822},
  urldate = {2024-05-11},
  langid = {english}
}

\end{document}